\documentclass{article}

\PassOptionsToPackage{numbers, compress, square}{natbib}

\usepackage[eandd, final]{neurips_2026}
\usepackage[utf8]{inputenc} % allow utf-8 input
\usepackage[T1]{fontenc}    % use 8-bit T1 fonts
\usepackage{hyperref}       % hyperlinks
\usepackage{url}            % simple URL typesetting
\usepackage{booktabs}       % professional-quality tables
\usepackage{amsfonts}       % blackboard math symbols
\usepackage{nicefrac}       % compact symbols for 1/2, etc.
\usepackage{microtype}      % microtypography
\usepackage[table]{xcolor}         % colors
\usepackage{pifont} 
\usepackage{multicol}
\usepackage{multirow}
\usepackage[export]{adjustbox}
\usepackage{pgfmath}
 \usepackage{tikz}
 \usepackage{amsmath}
 \usepackage{soul}
 \usepackage{gensymb}
 \usetikzlibrary{arrows.meta, positioning, fit, calc}
 \usepackage{subcaption}

\newcommand{\splg}{SP+LG}

\newcommand{\mAA}{mAA(10$^\circ$)}
\newcommand{\dsi}{$\delta_{1}$}
\newcommand{\dssi}{$\delta_{1}^{ai}$}
\newcommand{\relsi}{$\mathrm{Abs.Rel}$}
\newcommand{\relssi}{$\mathrm{Abs.Rel}^{ai}$}

\newcommand{\M}[1]{\mathbf{#1}}

\newcommand{\eth}{ETH3D}
\newcommand{\sintel}{Sintel}
\newcommand{\scannetpp}{ScanNet++}
\newcommand{\lamar}{LaMAR}

\newcommand{\baselinecalib}{$\mathsf{B}$}
\newcommand{\baselinesf}{$\mathsf{B}_{\mathrm{f}}$}
\newcommand{\calib}{$\mathsf{H}$}
\newcommand{\calibshift}{$\mathsf{H}_{\mathrm{a}}$}
\newcommand{\mysf}{$\mathsf{H}_{\mathrm{f}}$}
\newcommand{\sfshift}{$\mathsf{H}_{\mathrm{a,f}}$}

\newcommand{\mdecalib}{$\mathsf{K}$}
\newcommand{\mdecalibshift}{$\mathsf{K}_{\mathrm{a}}$}

\newcommand{\calibro}{$\mathsf{R}$}
\newcommand{\calibshiftro}{$\mathsf{R}_{\mathrm{a}}$}
\newcommand{\sfro}{$\mathsf{R}_{\mathrm{f}}$}
\newcommand{\sfshiftro}{$\mathsf{R}_{\mathrm{a,f}}$}

\newcommand{\mdecalibro}{$\mathsf{KR}$}
\newcommand{\mdecalibshiftro}{$\mathsf{KR}_{\mathrm{a}}$}

\newcommand{\ie}{\textit{i}.\textit{e}.}
\newcommand{\eg}{\textit{e}.\textit{g}.}

\newcommand{\rank}[3]{%
    \ifnum#2=0
        #1%
    \else
        \ifnum#3>1
            \edef\perf{\fpeval{100*(#2-1)/(#3-1)}}%
            \expanded{\noexpand\cellcolor{white!\perf!cyan}}#1%
        \else
            \cellcolor{white}#1%
        \fi
    \fi
}

\title{Depth2Pose: A Pose-Based Benchmark for Monocular Depth Estimation without Ground-Truth Depth}

\author{%
  Viktor Kocur$^{1}$\thanks{Equal contribution} \quad Sithu Aung$^{2\,\ast}$ \quad Gabrielle Flood$^{2\,\ast}$ \quad Yaqing Ding$^2$ \\\textbf{Lukas Bujnak}$^1$ \quad \textbf{Torsten Sattler}$^3$ \quad \textbf{Zuzana Kukelova}$^2$  \\
	$^1$ Faculty of Mathematics, Physics and Informatics, Comenius University in Bratislava\\
    $^2$ Visual Recognition Group, Faculty of Electrical Engineering, 
	Czech Technical University in Prague \\
    $^3$ Czech Institute of Informatics, Robotics and Cybernetics, Czech Technical University in Prague
}

\begin{document}

\maketitle

\begin{abstract}
Monocular depth estimation has improved significantly in recent years, driven by increasingly powerful models and large-scale training data. Predicted depth is increasingly used as an input signal for downstream tasks such as Structure-from-Motion (SfM), visual localization, and SLAM. However, monocular depth estimators (MDEs) are still primarily evaluated in terms of depth accuracy. Standard metrics aggregate errors globally and may not reflect the usefulness of depth for downstream geometric tasks.
We therefore propose Depth2Pose, a framework for evaluating MDEs in the context of downstream tasks. By combining depth predictions with feature correspondences in depth-aware geometric solvers, we use relative camera pose estimation accuracy as a task-driven proxy for depth quality. Traditional benchmarks require dense ground truth in the form of per-pixel depth, which is expensive to obtain. In contrast, our formulation requires only camera poses, which can be estimated efficiently, e.g., using Structure-from-Motion pipelines.
As a result, our framework can be applied to scenes where ground-truth depth is difficult to obtain, for example due to large scene scale or heavy occlusions (e.g., vegetated environments). Leveraging this, we introduce the D2P dataset, which contains challenging scenes outside the distribution of commonly used training data.
We show that methods performing well under standard depth error metrics on existing benchmarks also perform well under our pose-based metric when evaluated on the same datasets, but do not necessarily generalize to our more challenging dataset. Finally, we provide a simple and extensible evaluation framework. The dataset and code are available at \href{https://kocurvik.github.io/depth2pose/}{kocurvik.github.io/depth2pose}.
\end{abstract}

%----------------------------------------------------------------------------------
\section{Introduction}

Monocular depth estimation has become a central problem in computer vision, with rapid progress driven by deep learning and the availability of large-scale training data. Contemporary methods achieve strong performance on established benchmarks and are increasingly deployed as components within larger systems. In particular, predicted depth is starting to be used as an additional cue
in downstream tasks such as Structure-from-Motion (SfM)~\cite{liu2022depth, pataki2025mp, zhu2026marginalized}, 
visual localization~\cite{arnold2022map, brachmann2024scene}, %, jiang2026imloc}, 
Simultaneous Localization and Mapping (SLAM)~\cite{mur2017orb, teed2021droid, zhu2024nicer, Li2026DROIDW, dexheimer2024compact}, and dense 3D reconstruction~\cite{guo2022neural, yu2022monosdf, dong2023fast, han2025sparserecon}, where monocular depth maps help improve robustness and reduce ambiguities.

Despite their growing importance for downstream tasks, monocular depth estimation methods are still only evaluated in terms of their depth accuracy. 
In order to produce scalar metrics, existing evaluation protocols globally aggregate errors~\cite{eigen2014depth}. 
Common measures such as absolute relative error or root mean squared error aggregate discrepancies over all pixels and treat all regions equally. Threshold-based metrics (\eg, the percentage of pixels exceeding a given error) partially address this limitation by highlighting regions with large errors. However, they still do not explicitly account for the differing geometric importance of regions in downstream tasks.

Another limitation of existing and widely used benchmarks~\cite{geiger2013vision, silberman2012indoor, dai2017scannet, schops2017multi, vasiljevic2019diode, sun2020scalability, butler2012naturalistic,  li2018megadepth,tung2024megascenes,li2026long,yao2020blendedmvs,Richter_2017} lies in their dependence on per-pixel ground truth depth values. 
Obtaining this ground truth depth itself is challenging, restricting these benchmarks to scenes where specialized hardware (\eg, RGB-D cameras or LiDAR) can be employed, where multi-view stereo approaches can be used to obtain pseudo-ground truth~\cite{li2018megadepth,li2026long}, or to synthetic scenes~\cite{butler2012naturalistic,yao2020blendedmvs,Richter_2017}. 
As a consequence, many challenging scenarios, including large-scale outdoor scenes, reflective or transparent surfaces, and unconstrained environments, are underrepresented in current evaluation protocols.  
Furthermore, because ground-truth depth is typically used both for training and evaluation, benchmarks are often closely aligned with the training data distribution.
This overlap limits the assessment of generalization and makes it difficult to quantify performance in diverse, real-world environments. 

For many downstream geometric tasks, such as SfM or SLAM, it is not necessary for depth to be uniformly accurate across the entire image. Instead, it is often sufficient for depth to be highly precise in regions that are geometrically informative, \eg, areas with strong texture, stable feature correspondences, or favorable viewing geometry. Errors in less informative regions may have only a minor impact on final task performance. Consequently, two depth estimators with similar global or threshold-based error metrics can exhibit substantially different behavior when integrated into geometric pipelines. This discrepancy is largely invisible under standard evaluation protocols, which do not capture the task-dependent utility of depth.

At the same time, obtaining accurate camera poses is often significantly easier than acquiring dense ground-truth depth. Mature SfM %and SLAM 
pipelines can reliably reconstruct camera trajectories across a wide range of environments, including those where depth sensing is impractical. This observation motivates a shift in perspective: rather than evaluating monocular depth directly against ground truth, one can assess its quality indirectly through its impact on downstream geometric estimation.

In this work, we introduce \textbf{Depth2Pose}, a novel evaluation framework that measures the quality of monocular depth predictions via their contribution to relative pose estimation. Specifically, we employ modern depth-aware geometric solvers, such as those introduced in the 
RePoseD
framework~\cite{ding2025reposed},
which jointly leverage point correspondences and predicted depth to estimate relative camera motion.
Thus, Depth2Pose measures depth quality based on the usefulness of the predicted depth maps for accurate pose estimation. 
Importantly, this formulation does not require ground-truth depth and can therefore be applied to a much broader range of datasets. Moreover, by evaluating depth through its influence on geometric estimation, our framework naturally emphasizes regions of the scene that are most informative for many downstream tasks, allowing it to capture cases where depth is locally accurate and practically useful despite higher global error.

We validate our framework 
on several established benchmarks and show that pose-based evaluation correlates strongly with standard depth metrics, confirming its reliability as an %alternative 
evaluation signal for depth quality. 
In order to investigate the generalization capabilities of monocular depth estimators, 
we present the \textbf{D2P dataset}, which consists of challenging real-world scenes that lie outside the distribution of commonly used training and evaluation data. 
We show that good performance under depth- and pose-based metrics on standard depth accuracy benchmarks does not necessarily translate to good performance on our new dataset. 
As such, our dataset contributes to 
a more realistic assessment of monocular depth estimation methods. 
The result further validates our evaluation framework, as it can be used to easily create benchmark datasets that reflect real-world deployment scenarios without requiring per-pixel ground truth depth values. 

In summary, our contributions are as follows: 
    \textbf{(1)} we introduce \textbf{Depth2Pose}, a task-driven evaluation framework for monocular depth estimation based on depth-aware relative pose estimation.
    Depth2Pose represents a \emph{benchmarking paradigm} that replaces ground-truth depth with ground-truth poses, enabling evaluation in diverse real-world environments.
    \textbf{(2)} we validate our framework on 
    existing benchmarks for measuring depth accuracy. 
    Our results show that pose-based metrics correlate with standard depth metrics, 
    while additionally capturing task-relevant, spatially non-uniform depth quality. 
    \textbf{(3)} we present a new dataset of challenging real-world scenes beyond the distribution of existing benchmarks. 
    We show that good performance on standard benchmarks does not necessarily indicate good performance on our dataset, demonstrating that our dataset introduces challenges not present in existing benchmarks. 
    \textbf{(4)} 
    we provide a simple and extensible evaluation protocol to facilitate reproducible benchmarking. 

    Our dataset and code are publicly available at 
    \href{https://github.com/kocurvik/depth2pose}{github.com/kocurvik/depth2pose}.

%----------------------------------------------------------------------------------
\section{Related Work}
\noindent \textbf{Monocular depth estimation.}
Monocular depth estimation (MDE) has undergone rapid development with the emergence of deep learning methods.
Early approaches relied on supervised learning from RGB-D datasets~\cite{silberman2012indoor, geiger2013vision}, learning direct mappings from images to metric depth~\cite{eigen2014depth, laina2016deeper}.
More recent models leverage large-scale pretraining and heterogeneous training data to achieve strong cross-dataset performance and improved robustness~\cite{ranftl2021vision, ranftl2020towards, bhat2023zoedepth, guizilini2023towards, yin2023metric3d, wang2025moge, wang2025moge2, yang2024depth, piccinelli2025unidepthv2}.
In parallel, self-supervised methods~\cite{garg2016unsupervised, godard2019digging, watson2021temporal, yang2024depth1} reduce dependence on ground-truth depth supervision by exploiting photometric consistency and geometric constraints across multiple view~\cite{zhang2025survey}.

\noindent \textbf{Evaluation of Depth Estimation.}
Monocular depth estimation is predominantly evaluated using pixel-wise error metrics, including absolute relative error ($\rm{AbsRel}$), root mean squared error ($\rm{RMSE}$), and threshold-based accuracy ($\delta$)~\cite{eigen2014depth}.
These measures require dense ground-truth depth, restricting evaluation to datasets acquired with LiDAR or RGB-D sensors~\cite{silberman2012indoor, geiger2013vision, schops2017multi}, to scenes where multi-view stereo algorithms work well and can thus provide pseudo-ground truth~\cite{li2018megadepth,li2026long}, or to synthetic environments~\cite{yao2020blendedmvs, Richter_2017, ros2016synthia, cabon2020virtual, roberts2021hypersim}.
As a result, existing benchmarks are often closely aligned with training distributions, leaving open questions about how well the evaluated depth predictors generalize to out-of-distribution environments.
Furthermore, global aggregation of pixel-wise errors implicitly assumes uniform importance across image regions, which does not necessarily reflect
the geometric requirements of downstream tasks. 

Works that use monocular depth estimates as part of SfM~\cite{liu2022depth, pataki2025mp, zhu2026marginalized},
visual localization~\cite{arnold2022map, brachmann2024scene},  
SLAM~\cite{ teed2021droid, zhu2024nicer, Li2026DROIDW, dexheimer2024compact} or 3D detection~\cite{huang2024training, skvrna2025monosowa, lee2025plot} systems evaluate how different  depth estimators impact system performance.
Yet, we are not aware of prior work that analyzes the relation between per-pixel depth accuracy and downstream task performance, or the generalization of depth estimators to out-of-distribution scenes. 
We close this gap in the literature. 

\noindent \textbf{Depth-aware camera pose estimation.}
Recent work~\cite{ding2025reposed,barath2022relative,astermark2024fast,ding2024fundamental,yu2025relative} has introduced relative camera pose solvers that explicitly incorporate predicted depth, extending classical formulations based on epipolar geometry~\cite{Hartley_Zisserman_2004}.
These methods leverage per-point depth estimates together with image correspondences to better constrain camera motion, particularly in challenging or degenerate configurations.
Notably, the \emph{RePoseD} framework~\cite{ding2025reposed} proposes a family of robust depth-aware solvers designed to operate reliably under noisy and incomplete depth predictions.
In constrast to prior work~\cite{ding2025reposed,barath2022relative,astermark2024fast,ding2024fundamental,yu2025relative}, which primarily aims to improve pose estimation itself, we employ such depth-aware pose solvers as an evaluation mechanism.
By measuring the accuracy of relative poses recovered using predicted depth, we obtain a task-relevant proxy for depth quality that does not rely on ground-truth depth measurements. 
We chose relative pose estimation as the proxy task over other tasks such as SfM or visual localization due to two reasons: 
(1) the later lead to more complex pipelines that make it harder to isolate the impact of depth quality of individual images on system performance. 
(2) relative pose estimation is significantly more computationally efficient compared to complex pipelines. 

\noindent \textbf{Benchmark datasets.}
Established depth benchmarks, including KITTI~\cite{geiger2013vision} and NYU Depth~\cite{silberman2012indoor}, have enabled significant progress in monocular depth estimation but remain constrained by sensor-dependent data acquisition and limited environmental diversity.
More recent datasets~\cite{schops2017multi, chang2017matterport3d, sun2022neural, sarlin2022lamar, yeshwanth2023scannet++, wu2023omniobject3d, vasiljevic2019diode, sun2020scalability, xia2024rgbd} improve scale, scene diversity, or reconstruction quality; however, evaluation in these benchmarks still relies on ground-truth depth measurements.
In contrast, our approach enables depth evaluation in scenarios where only camera poses are available, substantially expanding the range of usable real-world data.
To this end, we introduce a new dataset of challenging real-world scenes that lie outside the distribution of commonly used training and evaluation benchmarks.

%----------------------------------------------------------------------------------
\section{Depth2Pose (D2P): Relative Pose-based Monodepth Evaluation}
\label{sec:depth2pose_framework}

\begin{figure}[t]
\centering
\includegraphics[width=0.92\linewidth]{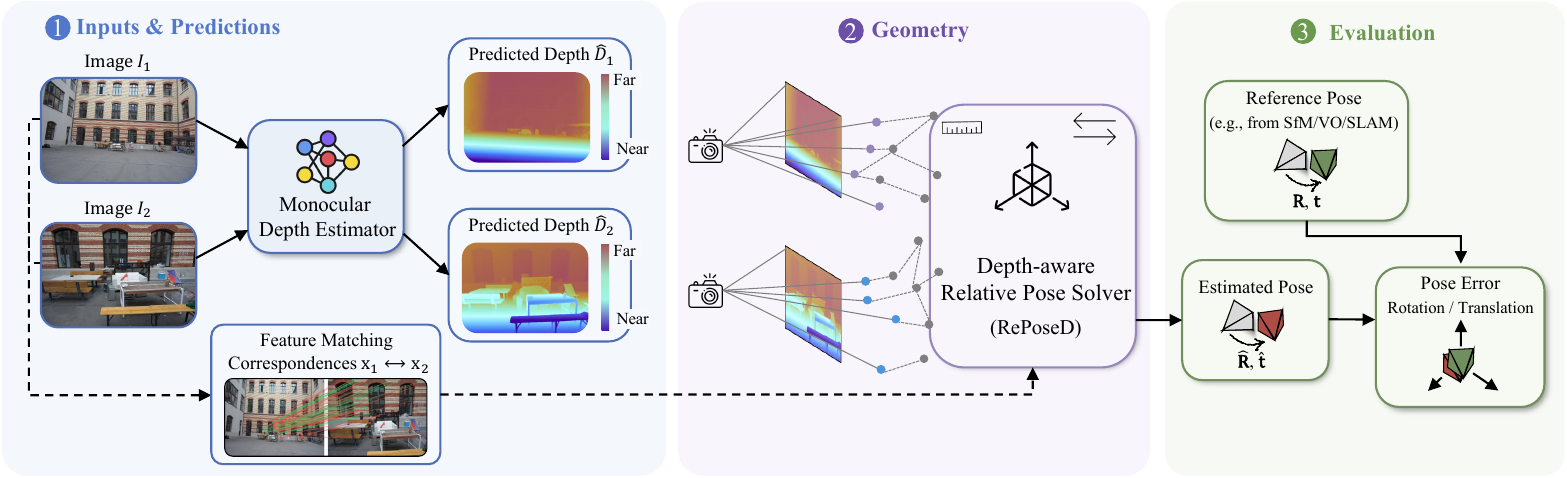} 
\caption{
Overview of our \textbf{Depth2Pose} evaluation framework: given a pair of visually overlapping images, we  estimate monocular depth estimates and use feature matching to obtain 2D-2D correspondences. The depth maps and matches are then used for depth-aware relative pose estimation~\cite{ding2025reposed}. As the accuracy of the estimated pose depends on the accuracy of the predicted depth maps, we measure depth map quality via the error between the estimated and ground truth relative pose.   
}
\label{fig:depth2pose_pipeline}
\end{figure}

Recent work
has shown that accurate monocular depth estimates can improve relative pose estimation when incorporated into geometric solvers~\cite{ding2025reposed,yu2025relative}. Based on this observation, we propose an evaluation methodology for monocular depth estimation (MDE) based on the principle that higher-quality depth predictions should lead to more accurate pose estimates when used within depth-aware solvers. We validate this assumption empirically in Sec.~\ref{sec:evaluation}. 

Our evaluation framework formulation enables the use of standard evaluation protocols for relative pose estimation~\cite{jin2021imc} as a proxy for assessing depth quality. A key advantage of this approach is that ground-truth camera poses are significantly easier to obtain than dense ground-truth depth, and can be reconstructed automatically from image collections using mature SfM pipelines such as COLMAP~\cite{schoenberger2016colmap}. Furthermore, by evaluating depth through its impact on pose estimation, our methodology directly reflects its utility in downstream geometric tasks.

\noindent \textbf{Benchmark philosophy.}
Our approach is guided by three 
principles:  (1) \emph{task relevance}: depth should be evaluated based on its utility for geometric reasoning, instantiated here through relative pose estimation. 
(2) \emph{scalability}: evaluation should not depend on specialized sensing hardware, enabling the use of diverse real-world datasets. 
(3) \emph{sensitivity to geometric structure}: evaluation should capture whether depth is accurate in regions that are important for downstream tasks, rather than treating all pixels uniformly.

By measuring the accuracy of the relative poses recovered using predicted depth, our framework satisfies these criteria. It naturally emphasizes geometrically informative regions, such as those supporting stable correspondences, and allows meaningful comparison of methods even when their global pixel-wise errors are similar. At the same time, it enables benchmarking in environments where traditional depth evaluation is not feasible.
We emphasize that our goal is not to replace standard depth metrics, but to complement them with a task-driven perspective.

\noindent \textbf{Evaluation pipeline.}
We consider individual scenes captured by multiple images. For each scene, we reconstruct camera poses using COLMAP~\cite{schoenberger2016colmap} (see Sec.~\ref{sec:dataset}). Following~\cite{jin2021imc}, we sample image pairs with at least 10\% visual overlap based on the SfM model.
As shown in Fig.~\ref{fig:depth2pose_pipeline}, for each selected image pair, we extract and match keypoints, and obtain depth values at the matched locations from the MDE under evaluation.\footnote{Matches with invalid depth estimates (NaN, $\infty$, negative, or zero values) are discarded.} The resulting set of correspondences augmented with depth is then used as input to depth-aware relative pose solvers. The estimated relative pose is compared to the reference pose obtained from the COLMAP reconstruction. 
We discuss individual steps 
in more detail below. 

\noindent \textbf{Keypoint detection and matching.}
We evaluate 
two approaches for feature extraction and matching. The first approach uses LoMa~\cite{nordstrom2026loma}, which resizes the longer image side to 1024 pixels while preserving aspect ratio. 
The second combines SuperPoint~\cite{detone2018superpoint} keypoints with the LightGlue~\cite{lindenberger2023lightglue} matcher (\splg), where keypoints are detected at the native image resolution. 
Both approaches use learned local features and learned feature matchers. 
We chose LoMa as a very recent state-of-the-art method while SP+LG is a commonly used baseline in the literature. 
For both methods, we retain at most 2048 correspondences per image pair. In the main paper, we only show results for LoMa, the results for SP + LG are shown in the appendix.

\noindent \textbf{Pose estimation.}
To estimate the relative pose from feature correspondences and depth, we use PoseLib~\cite{PoseLib}, which implements robust estimation using LO-RANSAC~\cite{lebeda2012fixing}. As a baseline, we consider the classical the 5-point solver~\cite{nister2004efficient} solver that relies only on point correspondences without depth. We denote this baseline as \baselinecalib{}.
For depth-aware pose estimation, we use the official PoseLib-based implementation of RePoseD~\cite{ding2025reposed}, which provides several solver variants with different assumptions on depth, including scale-invariant and affine-invariant formulations. 
As our primary configuration, we use the scale-invariant calibrated solver with a hybrid error, employing the Sampson error for scoring and a combination of Sampson and symmetric reprojection errors for local optimization. 
We denote this variant as \calib. This choice is motivated by its strong performance reported in~\cite{ding2025reposed}, as well as by the fact that many recent MDEs~\cite{wang2025moge2, piccinelli2025unidepthv2, piccinelli2025unik3d, lin2025dav3} provide metric depth.
Affine-invariant solvers can be beneficial in 
more challenging scenarios. We therefore include these variants in our evaluation and report their results in the appendix. These results are consistent with our main findings and do not change the overall conclusions. Additional details on the robust estimation procedure are also provided in the appendix.

Since the Sampson error depends only on image correspondences and not on depth, it is possible for inaccurate depth estimates to have limited influence on the final pose. To better isolate the contribution of depth, we additionally evaluate a variant that uses only the symmetric reprojection error for both scoring and local optimization. We denote this configuration as \calibro. This variant provides a more direct assessment of how depth accuracy impacts pose estimation.

Importantly, using both estimators \calib{} and \calibro{} for evaluation implicitly emphasizes geometrically informative regions, as only correspondences that contribute to stable pose estimation influence the final metric. While \calib{} leads to better poses, which provides practical insight into the performance of MDEs for pose estimation, \calibro{} typically results in less accurate poses, but provides a more direct assessment of 
the depth quality as the Sampson error can cover for inaccurate depth estimates. 
Additional solver variants, including uncalibrated settings, estimators assuming affine-invariant depth and configurations using camera intrinsics estimated by MDEs during inference, are evaluated in the appendix (Sec.~\ref{sec:more_results}) and summarized in Table~\ref{tab:estimators}.

%----------------------------------------------------------------------------------
\section{D2P Dataset for Pose-Focused Depth Evaluation}
\label{sec:dataset}

\begin{figure*}[!t]
  \centering
  \includegraphics[width=0.92\linewidth]{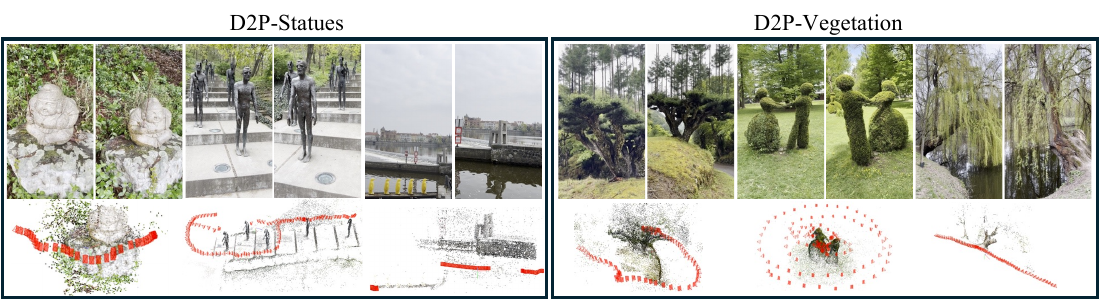}  
  \caption{Examples of the D2P dataset. The images (top) together with the corresponding camera poses (bottom) enable benchmarking MDEs without reference depth, using relative poses instead.}
  \label{fig:dataset_examples}
\end{figure*}

To enable evaluation of monocular depth estimation in diverse and challenging real-world scenarios, we introduce the \textbf{D2P dataset}. Unlike traditional depth benchmarks, our dataset is designed specifically for pose-based evaluation and does not require ground-truth depth.

\noindent \textbf{Dataset categories and design rationale.} 
The dataset consists of scenes grouped into two categories: D2P-Statues and D2P-Vegetation. 
D2P-Statues comprises sculptures located in public spaces, typically captured with full or near-complete $360^\circ$ coverage. These scenes exhibit predominantly %rigid 
geometry with well-defined structure, enabling reliable pose estimation while remaining challenging for monocular depth methods in terms of scale and shape interpretation.
All statues in the dataset are unique and do not appear in standard training datasets. 
Some statues are placed in non-easily accessible environments, such as being placed in water or being hung between buildings (\textit{cf.}  Fig.~\ref{fig:statues_overview} in the appendix). 
Moreover, several exhibit unconventional proportions, such as partial human figures or exaggerated anatomical features. 
At the same time, contextual cues from the surrounding environment are preserved, allowing for the inference of object scale.
Although such scenes could, in principle, be captured using active depth sensors, collecting RGB-only imagery enables significantly faster acquisition and facilitates outdoor data collection, where depth sensing is often less practical.

D2P-Vegetation includes scenes containing trees, shrubs, and park environments.
Some sequences are object-centric, whereas others are captured with dominant forward motion, reflecting realistic navigation scenarios.
We include both scenes %include vegetation both 
with and and scenes without foliage, resulting in fine-scale geometric detail as well as thin structures such as branches and leaves, which are known to challenge monocular depth estimation due to frequent depth discontinuities and partial occlusions.
Moreover, vegetation is difficult to capture reliably using active depth sensors due to sparsity, transparency effects, and complex geometry, and is therefore underrepresented in existing depth datasets. 
This category is designed to evaluate the robustness of depth predictions in highly structured yet irregular natural environments.

In total, our D2P dataset contains 24 scenes, with 12 scenes per subset (D2P-Statues and D2P-Vegetation) and 1,942 images in total.
Fig.~\ref{fig:dataset_examples} shows examples. All scenes are shown in 
Sec.~\ref{sec:d2p_appendix}.

\noindent \textbf{Capturing devices and conditions.} 
Images are captured using RGB cameras from multiple consumer devices, including Google Pixel 9, iPhone 14, 15 pro, 16 pro, and a GoPro camera.
During recording, the focal length was kept fixed within each sequence to ensure consistent camera intrinsics for pose estimation.
Videos were acquired using slow handheld motion either around a central object or through a scene.
Although handheld capture results predominantly in horizontal motion, camera trajectories still span the three-dimensional space, providing sufficient viewpoint diversity.

Frames were sampled at 1 FPS to reduce redundancy while preserving meaningful viewpoint changes.
A small number of frames were removed when strong occlusions, motion blur, or similar artifacts were present, in order to avoid evaluation failures unrelated to depth estimation quality.
Each scene was recorded within a short time interval, resulting in largely consistent background appearance and lighting conditions across individual sequences, with only minor exceptions.

\noindent \textbf{Pseudo ground truth generation.} 
We generate camera pose estimates serving as pseudo ground truth (pGT) using COLMAP~\cite{schoenberger2016colmap,schoenberger2016mvs}, a widely adopted Structure-from-Motion framework that is also commonly used as a pseudo ground truth generator.
Reconstruction was performed using a radial distortion camera model with a shared focal length across images within each scene. 
For evaluation we use undistorted images.
Feature extraction and matching were adapated to scene characteristics: for vegetation scenes, we employed ALIKED~\cite{zhao2023aliked} features combined with the LightGLUE~\cite{lindenberger2023lightglue} matcher, while SIFT~\cite{lowe2004distinctive} features with brute-force matching were used for the statues subset.
We tried to ensure the quality of the estimated poses by enforcing 
that reconstructions have a reprojection error below one pixel, removing blurred images or images with high reprojection errors otherwise.  
Only successfully reconstructed images were retained for evaluation.
To ensure privacy, all images were anonymized by automatically masking people using the easy-anno tool~\cite{easyanon}.

%----------------------------------------------------------------------------------
\section{Evaluation}
\label{sec:evaluation}
In this section, we present benchmarking results obtained using our pose-based evaluation protocol.
We first introduce the evaluation metrics, and describe baselines and implementation details in Sec.~\ref{sec:implementation}.
Sec.~\ref{sec:validation} then investigates the relation between relative pose accuracy 
and performance under classical depth accuracy metrics on established monocular depth estimation benchmarks. 
Finally, Sec.~\ref{sec:d2p_eval} presents evaluation results on our D2P dataset.

%-------------------------------------------------------
\subsection{Evaluation Setup}
\label{sec:implementation}
\noindent \textbf{Depth metrics.}
To enable comparison with different monocular depth estimation models, we evaluate predicted depth using standard scale-invariant and affine-invariant metrics~\cite{eigen2014depth}.
Specifically, we report the absolute relative error ($\mathrm{Abs.Rel} \downarrow$), \ie, $| z_{gt} - z_{est}| / z_{gt}$, and the threshold accuracy ($\delta_1\uparrow$), defined as the percentage of pixels satisfying $\mathrm{max}(z_{gt}/z_{est}, z_{est}/z_{gt}) < 1.25$, where subsrcipts $gt$ and $est$ denote ground-truth and estimated values.
Metrics are computed over dense depth maps while excluding pixels without valid ground-truth measurements.

\noindent \textbf{Pose metrics.}
We follow \cite{jin2021imc} and calculate the rotation error and translation error as:
\begin{equation}
    e_{\M R} = \textnormal{arccos} \left(\frac{\textnormal{Tr} \left(\M R_{est}^\top \M R_{gt} \right) - 1}{2}\right), \qquad
    e_{\M{t}} = \textnormal{arccos}\left( \frac{\M{t}_{est} \cdot \M{t}_{gt}}{ || \M{t}_{est} || \cdot||\M{t}_{gt}||}  \right) \enspace .
\end{equation}
The overall pose error is defined as $e_p = \textnormal{max}\left(e_{\M R}, e_{\M{t}} \right)$~\cite{jin2021imc}.
Following standard relative pose estimation benchmark practice~\cite{jin2021imc}, performance is summarized using the mean Average Accuracy (mAA), computed as the area under the normalized cumulative distribution of pose errors up to a threshold of 10\degree. 
This metric is denoted as \mAA$\uparrow$.

%-------------------------------------------------------
\noindent \textbf{Baselines.}
We evaluate a diverse set of monocular depth estimation models, including UniDepthV1~\cite{piccinelli2024unidepth} and V2~\cite{ piccinelli2025unidepthv2}, DepthPro~\cite{bochkovskii2024depth}, MoGeV1~\cite{wang2025moge} and V2~\cite{ wang2025moge2}, DepthAnythingV2~\cite{yang2024depth}, UniK3D~\cite{piccinelli2025unik3d}, Metric3DV2~\cite{yin2023metric3d} and  InfiniDepth~\cite{yu2026infinidepth}.
In addition, we include feed-forward reconstruction approaches: 
Pi3~\cite{wang2025pi3}, MapAnything~\cite{keetha2025mapanything} and DepthAnythingV3~\cite{lin2025dav3}, which we evaluate under a monocular setting using a single input image.
For all methods, inference is performed using the offical implementations released by the respective authors with all available backbones.
Images are resized according to each model's recommended preprocessing pipeline, and predicted depth maps are upsampled back to the original resolution when required.
Depth values at feature locations are obtained using nearest-neighbor sampling.
Where supported, we evaluate models both with and without camera intrinsics provided at inference time.

\noindent \textbf{Hyperparameters.}
Following \cite{ding2025reposed}, we set the Sampson error threshold to 2 pixels and the reprojection error threshold to 16 pixels, and run RANSAC for 1,000 iterations.
Details about the image pairs selection on each dataset is mentioned in the appendix Sec.\ref{sec:standard_benchmark_info}.

%-------------------------------------------------------
\subsection{Comparison to Standard Depth Metrics}
\label{sec:validation}

\begin{figure}
    \centering
\scriptsize
\begin{tabular}{*{6}{c}}        
    \includegraphics[width=12pt, valign=m]{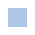} UniDepth1 
    & \includegraphics[width=12pt, valign=m]{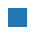} UniDepth2     
    & \includegraphics[width=12pt, valign=m]{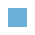} UniK3D   
    & \includegraphics[width=12pt, valign=m]{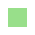} MoGeV1 
    & \includegraphics[width=12pt, valign=m]{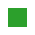} MoGeV2   
    & \includegraphics[width=12pt, valign=m]{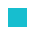} Metric3DV2 
    
\end{tabular}
\vspace{-1ex}

\begin{tabular}{*{4}{c}}    
    \includegraphics[width=12pt, valign=m]{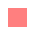} DAv3-Mono 
    & \includegraphics[width=12pt, valign=m]{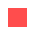} DAv3-Metric     
    & \includegraphics[width=12pt, valign=m]{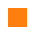} Pi3 
    & \includegraphics[width=12pt, valign=m]{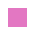} MapAnything
\end{tabular}
\vspace{-1ex}

\begin{tabular}{rccccc}
Backbone:
& \includegraphics[width=12pt, valign=m]{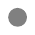} ViT-S
& \includegraphics[width=12pt, valign=m]{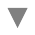} ViT-B
& \includegraphics[width=12pt, valign=m]{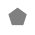} ViT-L
& \includegraphics[width=12pt, valign=m]{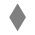} ViT-G
& \includegraphics[width=12pt, valign=m]{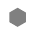} ConvNext
\end{tabular}
\vspace{-1ex}

\begin{tabular}{rcc}
Uses Intrinsics at Inference: 
& \includegraphics[width=12pt, valign=m]{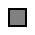} Yes
& \includegraphics[width=12pt, valign=m]{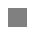} No
\end{tabular}

\includegraphics[width=0.43\linewidth]{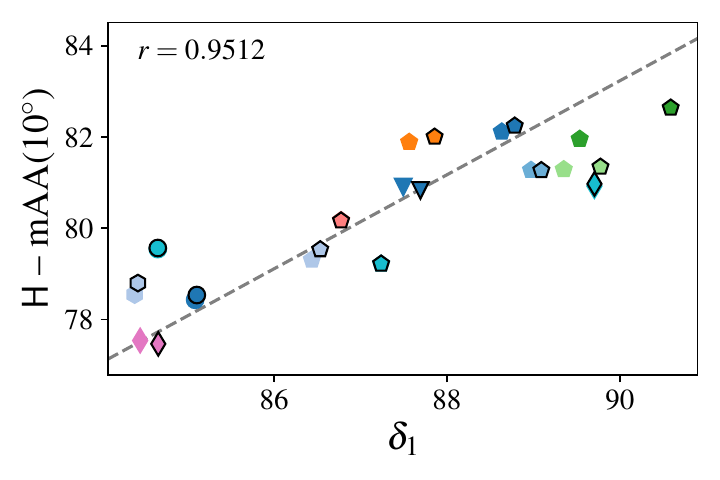} \hfill
\includegraphics[width=0.43\linewidth]{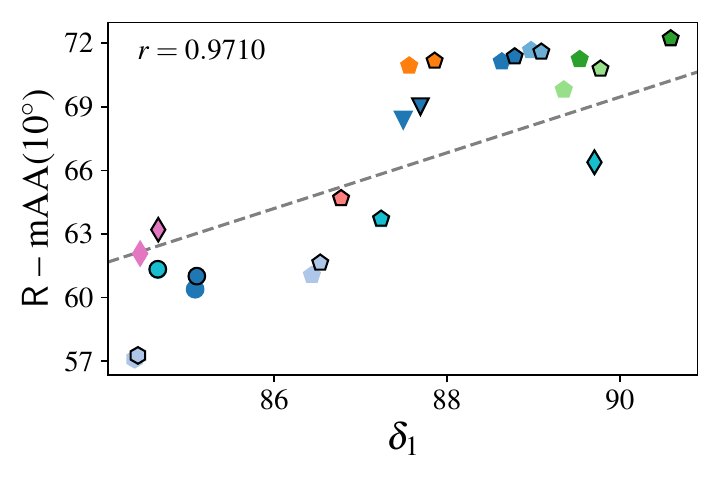} \vspace{-3ex}
    \caption{The values of \dsi{} compared to \mAA{} obtained using the \calib{} and \calibro{} estimators on the standard benchmark datsets \cite{schops2017multi,sarlin2022lamar,yeshwanth2023scannet++,butler2012naturalistic} using LoMa point correspondences~\cite{nordstrom2026loma} and different evaluted MDEs. The dashed line represents the linear fit of the data (the Pearsson correlation coefficient $r$ is provided in the plots). Results for DAv2 and DepthPro are outside of the shown range.} % (see Fig.~\ref{fig:corr_plot_sm} in the Appendix for full plot).}
    \label{fig:corr_plot}
\end{figure}

As a first experiment, we compare performance under the relative pose accuracy metric (used by our evaluation framework) with performance under classical depth accuracy metrics (\dsi{} and \relsi{}) on four standard datasets: \eth{}~\cite{schops2017multi}, \scannetpp{}~\cite{yeshwanth2023scannet++}, \sintel{}~\cite{butler2012naturalistic}, and \lamar{}~\cite{sarlin2022lamar} (see Sec.~\ref{sec:standard_benchmark_info} of the appendix for details on the evaluation setup). 
Fig.~\ref{fig:corr_plot} shows the results as well as the linear fit along with its Pearson correlation coefficient. %These results show that pose \mAA{} is highly correlated with \dsi. 
We further show results for the \relsi{}, \dsi{}, and \mAA{} metrics in Table~\ref{tab:main_table}. 
To obtain a single value, we first compute the mean per dataset (averaged over all images in a dataset) and then the mean of the per-dataset means.

Tab.~\ref{tab:main_table} shows that 
the \relsi{} metric is not strongly correlated with \dsi{} and \mAA{}. 
This is expected, as the latter two metrics measure the proportion of accurate depth with respect to some thresholds, and errors beyond these thresholds do not influence them.

\begin{table}[]
    \centering
    \caption{Evaluation results based on standard metrics (\dsi, \relsi) for standard benchmarks  \cite{schops2017multi,sarlin2022lamar,yeshwanth2023scannet++,butler2012naturalistic} and the \mAA{} of the estimated poses for the standard benchmarks and our D2P dataset. 
    For each MDE we show only the best performing variant (backbone, using known intrinsics at inference) based on the \dsi metric. Results for all variants are presented in Sec.~\ref{sec:more_results} of the appendix. Cell color denotes the rank of the method in the respective columns.}
    \label{tab:main_table}
\resizebox{\linewidth}{!}{

\begin{tabular}{lccccccccccc}
\toprule
  &   &   &   & \multicolumn{8}{c}{\mAA} \\ \cmidrule(lr){5-12}
  & MDE & \multirow{2}{*}{\dsi} & \multirow{2}{*}{\relsi} & \multicolumn{2}{c}{Standard} & \multicolumn{2}{c}{D2P-Mean} & \multicolumn{2}{c}{D2P-Statues} & \multicolumn{2}{c}{D2P-Vegetation} \\ \cmidrule(lr){5-6} \cmidrule(lr){7-8} \cmidrule(lr){9-10} \cmidrule(lr){11-12}
MDE-Backbone & w/$\M K$ &   &   & \calib{} & \calibro{} & \calib{} & \calibro{} & \calib{} & \calibro{} & \calib{} & \calibro{} \\ \midrule
MoGeV2-L & \checkmark & \rank{90.59}{1}{12} & \phantom{1}\phantom{1}\rank{9.45}{1}{12} & \rank{82.65}{1}{12} & \rank{72.22}{1}{12} & \rank{75.02}{7}{12} & \rank{59.70}{5}{12} & \rank{83.66}{3}{12} & \rank{71.82}{3}{12} & \rank{72.93}{10}{12} & \rank{48.66}{7}{12} \\
MoGeV1-L & \checkmark & \rank{89.78}{2}{12} & \phantom{1}\rank{10.47}{4}{12} & \rank{81.35}{4}{12} & \rank{70.79}{5}{12} & \rank{76.62}{3}{12} & \rank{62.23}{2}{12} & \rank{86.39}{1}{12} & \rank{76.18}{1}{12} & \rank{76.63}{6}{12} & \rank{53.38}{4}{12} \\
Metric3DV2-G & \checkmark & \rank{89.71}{3}{12} & \phantom{1}\rank{10.34}{3}{12} & \rank{80.97}{6}{12} & \rank{66.38}{6}{12} & \rank{74.32}{9}{12} & \rank{50.31}{9}{12} & \rank{82.15}{5}{12} & \rank{57.96}{9}{12} & \rank{71.84}{11}{12} & \rank{40.80}{12}{12} \\
UniK3D-L & \checkmark & \rank{89.09}{4}{12} & \phantom{1}\rank{14.37}{8}{12} & \rank{81.27}{5}{12} & \rank{71.59}{2}{12} & \rank{76.81}{2}{12} & \rank{61.94}{3}{12} & \rank{82.72}{4}{12} & \rank{67.94}{4}{12} & \rank{79.95}{1}{12} & \rank{60.30}{1}{12} \\
UniDepth2-L & \checkmark & \rank{88.78}{5}{12} & \phantom{1}\rank{12.88}{6}{12} & \rank{82.25}{2}{12} & \rank{71.37}{3}{12} & \rank{76.47}{4}{12} & \rank{59.91}{4}{12} & \rank{81.80}{6}{12} & \rank{64.83}{6}{12} & \rank{79.57}{2}{12} & \rank{58.91}{2}{12} \\
Pi3-L & \checkmark & \rank{87.86}{6}{12} & \phantom{1}\rank{10.24}{2}{12} & \rank{82.01}{3}{12} & \rank{71.16}{4}{12} & \rank{78.26}{1}{12} & \rank{64.45}{1}{12} & \rank{86.27}{2}{12} & \rank{75.28}{2}{12} & \rank{76.50}{7}{12} & \rank{51.85}{6}{12} \\
DAv3-Metric-L &  & \rank{86.78}{7}{12} & \phantom{1}\rank{12.26}{5}{12} & \rank{80.17}{7}{12} & \rank{64.68}{7}{12} & \rank{72.74}{10}{12} & \rank{50.90}{8}{12} & \rank{79.22}{9}{12} & \rank{58.54}{8}{12} & \rank{75.66}{8}{12} & \rank{47.82}{9}{12} \\
UniDepth1-L & \checkmark & \rank{86.53}{8}{12} & \phantom{1}\rank{13.31}{7}{12} & \rank{79.53}{9}{12} & \rank{61.64}{10}{12} & \rank{75.07}{6}{12} & \rank{49.96}{10}{12} & \rank{80.93}{8}{12} & \rank{54.22}{10}{12} & \rank{77.35}{5}{12} & \rank{47.84}{8}{12} \\
DepthPro-L &  & \rank{86.36}{9}{12} & \rank{208.04}{12}{12} & \rank{79.66}{8}{12} & \rank{61.85}{9}{12} & \rank{74.86}{8}{12} & \rank{56.39}{7}{12} & \rank{76.91}{10}{12} & \rank{60.57}{7}{12} & \rank{79.09}{3}{12} & \rank{52.33}{5}{12} \\
MapAnything-G & \checkmark & \rank{84.66}{10}{12} & \phantom{1}\rank{15.87}{9}{12} & \rank{77.46}{10}{12} & \rank{63.20}{8}{12} & \rank{75.15}{5}{12} & \rank{58.79}{6}{12} & \rank{81.31}{7}{12} & \rank{67.89}{5}{12} & \rank{77.43}{4}{12} & \rank{54.06}{3}{12} \\
DAv3-Mono-L &  & \rank{73.56}{11}{12} & \phantom{1}\rank{19.03}{10}{12} & \rank{70.33}{11}{12} & \rank{49.07}{11}{12} & \rank{50.21}{11}{12} & \rank{26.06}{11}{12} & \rank{54.35}{11}{12} & \rank{25.82}{11}{12} & \rank{74.98}{9}{12} & \rank{41.24}{10}{12} \\
DepthAnythingV2-B &  & \rank{60.53}{12}{12} & \rank{102.49}{11}{12} & \rank{56.48}{12}{12} & \rank{30.61}{12}{12} & \rank{46.22}{12}{12} & \rank{24.05}{12}{12} & \rank{51.53}{12}{12} & \rank{23.98}{12}{12} & \rank{69.28}{12}{12} & \rank{41.10}{11}{12} \\
\midrule GT &  &   &   & \rank{90.70}{0}{12} & \rank{93.73}{0}{12} & \rank{93.39}{0}{12} & \rank{97.65}{0}{12} & \rank{95.80}{0}{12} & \rank{98.87}{0}{12} & \rank{94.53}{0}{12} & \rank{97.69}{0}{12} \\
\midrule No Depth + \baselinecalib{} &  &   &   & \multicolumn{2}{c}{84.01} & \multicolumn{2}{c}{82.34} & \multicolumn{2}{c}{88.78} & \multicolumn{2}{c}{86.31} \\
\bottomrule
\end{tabular}
}
\end{table}

\begin{figure*}
  \centering
  \includegraphics[width=\linewidth]{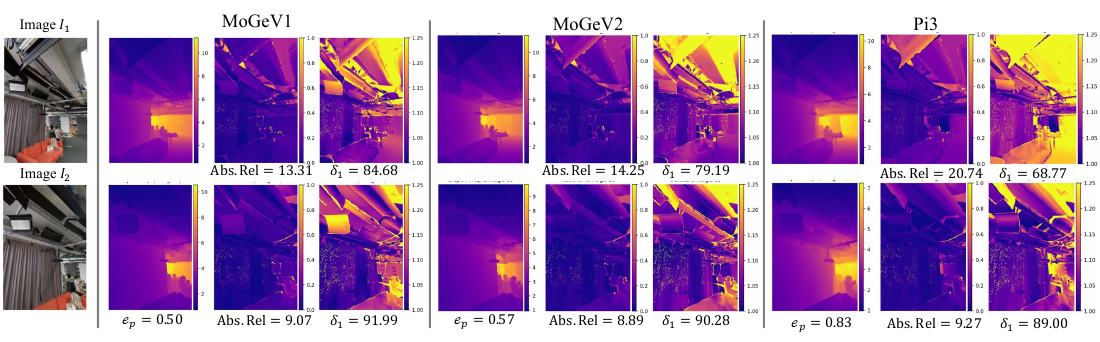}  
  \caption{Qualitative comparison of an image pair from the \lamar{} dataset. For each MDE, we show the predicted depth map together with error maps under the \relsi{} and \dsi{} metrics. For \dsi{}, all regions with a maximum ratio of 1.25 or larger are color-coded in yellow. $e_p$ denotes the relative pose error using our metric. As can be seen, the pose error $e_p$ is not directly correlated to \dsi{}.}
  \label{fig:lamar_sample1}
\end{figure*}

Fig.~\ref{fig:corr_plot} shows that, at least on a global level, there is a correlation between the \dsi{} and \mAA{} metrics. 
Similarly, Tab.~\ref{tab:main_table} shows that methods that are ranked high under the \dsi{} metric often also ranked high under the \mAA{} metric.  
This observation makes sense as more pixels with depth errors under a threshold (higher \dsi{}) should increase the chances for accurate pose estimation. 
However, we also note that from a certain value of \dsi{} onwards, the benefit of having more "good" pixels does not improve \mAA{}. For example, Pi3-L is measurably worse than MoGeV2-L under the \dsi{} metric while both perform similarly well under the \mAA{} metric. 
This behavior can be explained by the fact that it can be sufficient for accurate pose estimation to have accurate depth estimates in a few regions of the image. 
This is illustrated in Fig.~\ref{fig:lamar_sample1}, which shows an image pair for which MDEs achieve different \dsi{} and \relsi{} metrics while resulting in similar pose accuracy. This occurs due to the depth errors mostly accumulating in areas of low geometrical significance (\eg,  textureless ceilings). 
Additional examples for this behavior are shown in the appendix Fig.~\ref{fig:lamar_samples_appendix}. 

Fig.~\ref{fig:lamar_sample1} shows that there is not necessarily a correlation between the \dsi{} and \mAA{} metrics. 
We attribute the correlation apparently observed in Fig.~\ref{fig:corr_plot} and Tab.~\ref{tab:main_table} to the nature of the standard benchmarks: 
the standard benchmarks are quite similar to the training data of the MDEs. 
Thus, most MDEs perform quite well on most images, leading to high \dsi{} scores. 
As argued above, a high \dsi{} score is not necessary for accurate pose estimation. 
As there are basically no MDEs with lower \dsi{} scores on the standard dataset, it is hard to observe that the \dsi{} and \mAA{} metrics are not correlated. 

%-------------------------------------------------------
\begin{figure}[t]
\centering
\includegraphics[width=0.98\linewidth]{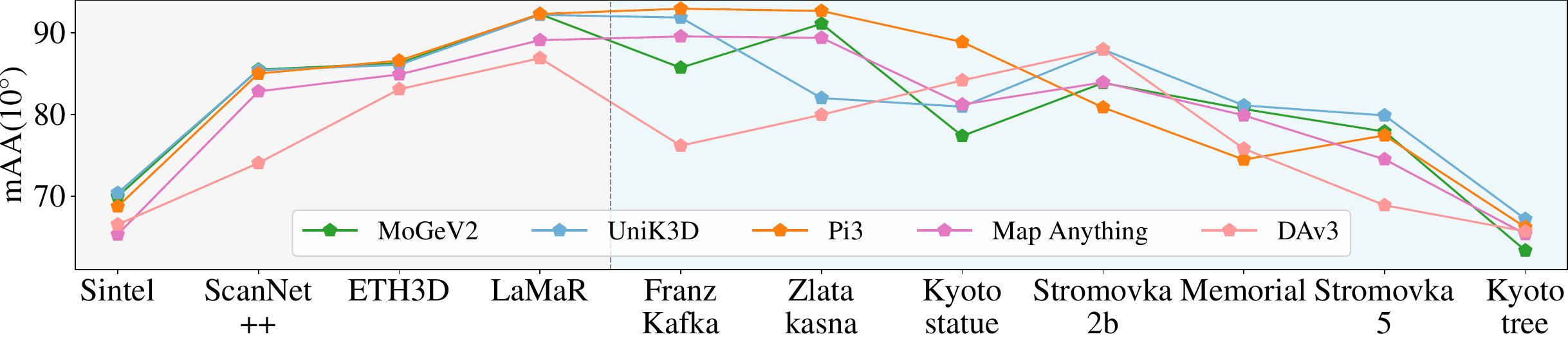} 
\caption{
Performance in terms of \mAA{} using the \calib{} estimator for a subset of the evaluated MDEs on the standard datasets and a subset of the scenes from the D2P dataset. The MDEs are chosen based on their rank on the standard datasets. All MDEs use known intrinsics during inference. 
}
\label{fig:d2p_ranks}
\end{figure}

\subsection{Evaluation on the D2P Dataset}
\label{sec:d2p_eval}
In our next experiment, we show that our D2P dataset contains challenging scenarios that are not covered by standard benchmarks. 
To this end, Fig.~\ref{fig:d2p_ranks} compares \mAA{} performance of selected MDEs on the standard benchmarks and selected scenes from the D2P dataset. 
We selected MDEs (MoGeV2, Pi3, UniK3D) that perform well in terms of \mAA{} (and also in terms of \dsi{}) when using the \calib{} on the standard benchmarks, as well as MDEs that perform less well (DepthAnythingV3, MapAnything).  
Fig.~\ref{fig:d2p_ranks} shows a consistent relative ranking between the methods among all four standard benchmarks. 
However, on scenes from our datasets, the rankings vary strongly, with the worst method (DepthAnythingV3) on standard benchmarks achieving the (joint) best rank on one scene and the best method (MoGeV2) falling down to the last rank on at least one scene. 
Notice the large changes in \mAA{} on our scenes. 

Tab.~\ref{tab:main_table} shows results aggregated over the full D2P dataset and over its two categories (for extended results, see Sec.~\ref{sec:more_results}). 
The table reflects the results from Fig.~\ref{fig:d2p_ranks}, namely that the methods that perform best / worst on standard benchmarks do not perform best / worst on our dataset. 
For example, the best-performing method under the \mAA{} metric on the standard benchmarks (MoGeV2) performs worse with the \calib{} estimator on D2P-Vegetation than %DepthAnythingV3 
DAv3-Mono-L
(the second-to-worst method on the standard benchmarks). 
Overall, our results show that both our evaluation protocol and our dataset provide additional insights into the behavior of MDEs that cannot be derived from the performance on standard benchmarks under standard depth accuracy metrics. 

In order to understand the \mAA{} performance obtained with the different MDEs on both the standard benchmarks and our dataset, we compare to two baselines in Tab.~\ref{tab:main_table}. 
\textit{No Depth + \baselinecalib{}} simply ignores depth values and only uses 2D-2D correspondences for relative pose estimation. 
As can be seen, using MDEs does not (yet) lead to improvements in pose accuracy on the datasets considered in this work.\footnote{\cite{ding2025reposed,yu2025relative} showed improvement over this baselines on the datasets used in their work for experimental evaluation. Our results show that current MDEs are not yet widely applicable for improving relative pose accuracy.} 
The \textit{GT} baseline illustrates the performance that could be achieved with better MDEs: 
It uses the known relative poses and the 2D-2D correspondences to triangulate a sparse set of 3D points. 
The depths of these points are then used for pose estimation. 
As can be seen, there is significant potential for improvement, \ie, our Depth2Pose framework and D2P dataset offers an avenue for measuring progress in MDE quality.

%----------------------------------------------------------------------------------
\section{Conclusion}
\label{sec:conclusion}
We introduced \textbf{Depth2Pose}, a framework for evaluating MDEs through their impact on relative pose estimation. By leveraging depth-aware geometric solvers, our approach provides a task-driven alternative to traditional pixel-wise metrics and removes the need for ground-truth depth, enabling scalable evaluation in diverse real-world environments.
We showed that pose-based evaluation provides new insights into the behavior of MDEs that cannot be observed from standard metrics 
on existing benchmarks. 
We introduced the \textbf{D2P dataset}, which extends evaluation to scenes with complex geometry, occlusions, and large depth variation.
Comparing results on standard benchmarks and our dataset, we revealed a generalization gap. 
In particular, methods performing well on standard datasets do not necessarily maintain performance on our more challenging scenes.
Together with a simple and extensible evaluation pipeline, our framework enables benchmarking in previously inaccessible scenarios.
We hope this work encourages a shift toward task-driven evaluation of MDEs. 

\noindent \textbf{Limitations.}  
Our framework is only applicable to scenes where accurate camera poses can be obtained, \eg, via SfM. 
Due to the scene scale ambiguity of relative pose estimation, it is not possible to estimate the metric accuracy of the estimated depth. 
More complex tasks (SfM, SLAM) can better compensate for individual inaccurate depth maps, \eg, redundancy between video frames reduces the impact of individual depth maps with high errors  on the overall system. 
Such systems can thus tolerate larger errors in the depth maps compared to the task of relative pose estimation. 
Our framework might thus underestimate the usefulness of some MDEs
for these tasks. 
Our framework only considers regions in the images where correspondences can be estimated. 
Thus, we cannot distinguish between an MDE that predicts accurate depth values everywhere, including in regions where no matches are found, \eg, on shiny objects such as mirrors, 
and an MDE that produces accurate depth estimates only in regions where correspondences are found.

%----------------------------------------------------------------------------------
%\section*{Acknowledgements}
%This work was supported by the Czech Science Foundation (GACR) JUNIOR STAR, Grant No.~22-23183M.
%The access to the computational infrastructure of the OP VVV funded project %CZ.02.1.01/0.0/0.0/16\_019/0000765 ``Research Center for Informatics'' is also gratefully acknowledged.

\medskip

%----------------------------------------------------------------------
\bibliographystyle{ieeetr}
\bibliography{mdrpbench_neurips26}

%%%%%%%%%%%%%%%%%%%%%%%%%%%%%%%%%%%%%%%%%%%%%%%%%%%%%%%%%%%%
\newpage

\appendix

\section{Details of the Experiments}

\subsection{Standard Benchmark}
\label{sec:standard_benchmark_info}
For the standard benchmarks, we use \eth~\cite{schops2017multi}, \lamar~\cite{sarlin2022lamar}, \scannetpp~\cite{yeshwanth2023scannet++}, and \sintel~\cite{butler2012naturalistic}. For the evaluation of the standard depth metrics \dsi{} and \relsi{}, we use the code from~\cite{wang2025moge2}. To estimate pose, we sample all valid pairs (10\% overlap criterion~\cite{jin2021imc}) from the ETH3D multiview dataset. For \lamar, we use only the data from the Navvis cameras due to the high quality of the associated GT depth. For all three scenes, we use only cam\_1 and sample 2500 valid image pairs for each scene. For Sintel, we remove the scenes ambush\_6, cave\_2, market\_2,  market\_6, temple\_3 from evaluation due to insufficient or degenerate motions following \cite{zhao2022particlesfm}. We sample 250 valid pairs for each of the remaining scenes. For \scannetpp{}, we sample 500 valid image pairs per scene.

\subsection{D2P Dataset}
\label{sec:d2p_appendix}

\begin{figure*}[!t]
  \centering
  \includegraphics[width=\linewidth]{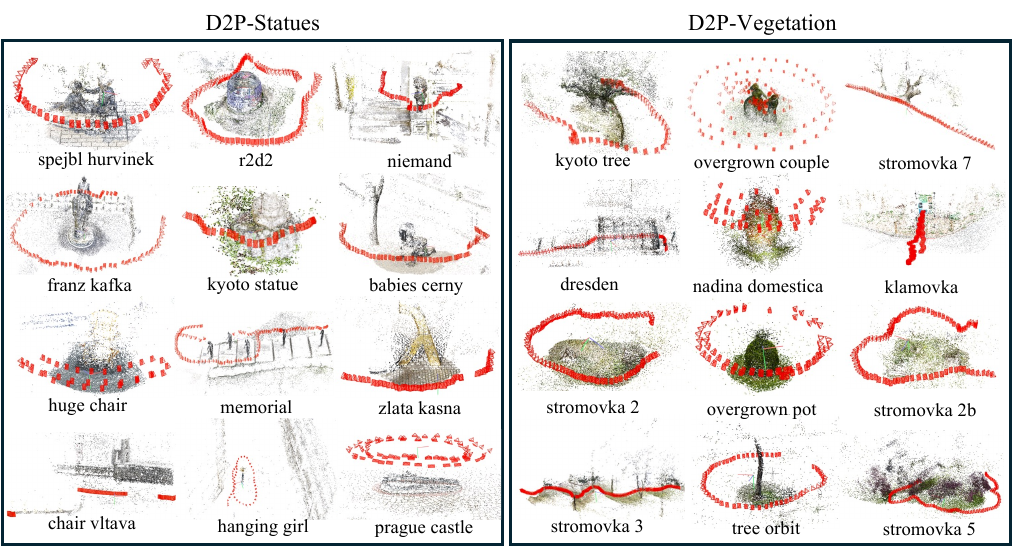}  
  \caption{COLMAP reconstructions of the scenes in our D2P dataset.}
  \label{fig:d2p_colmap}
\end{figure*}

\begin{figure*}[t]
    \centering
    \begin{tabular}{cccccc}
        \includegraphics[width=0.15\linewidth]{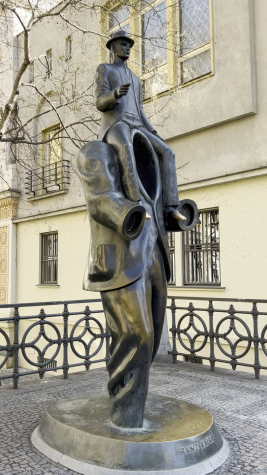} &
        \includegraphics[width=0.15\linewidth]{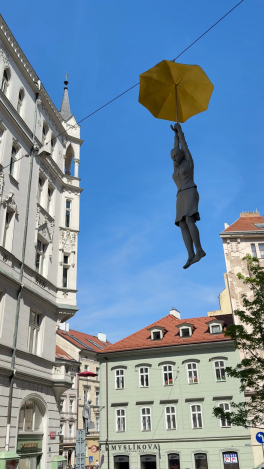} &
        \includegraphics[width=0.15\linewidth]{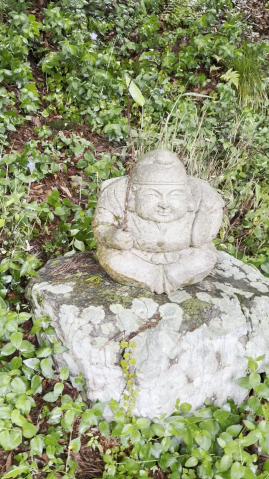} &
        \includegraphics[width=0.15\linewidth]{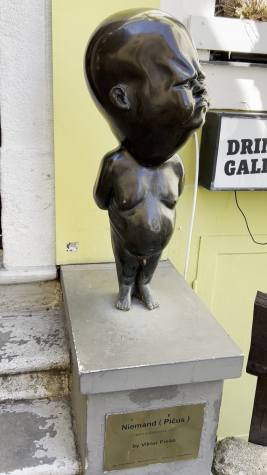} &
        \includegraphics[width=0.15\linewidth]{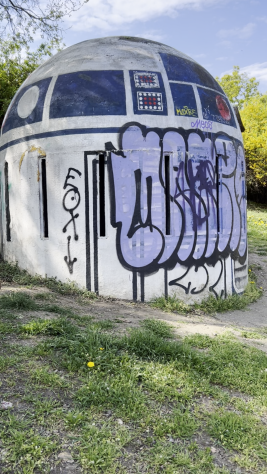} &
        \includegraphics[width=0.15\linewidth]{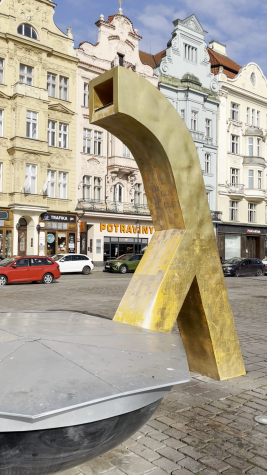} \\
        franz kafka & hanging girl & kyoto statue & niemand & r2d2 & zlata kasna \\
    \end{tabular}

    \vspace{1em}

    \begin{tabular}{cccccc}
        \includegraphics[width=0.15\linewidth]{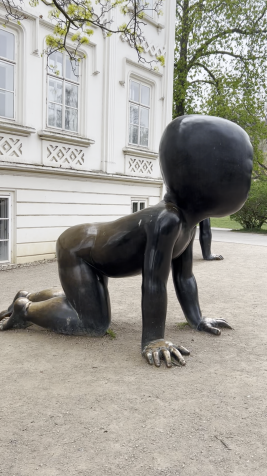} &
        \includegraphics[width=0.15\linewidth]{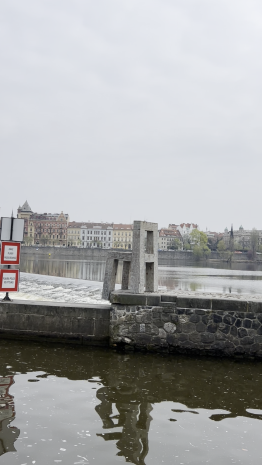} &
        \includegraphics[width=0.15\linewidth]{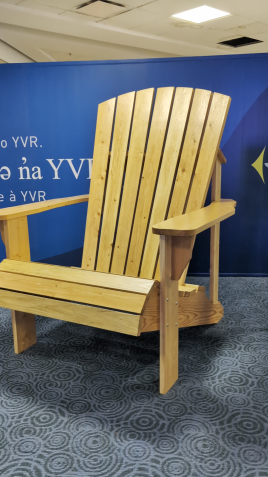} &
        \includegraphics[width=0.15\linewidth]{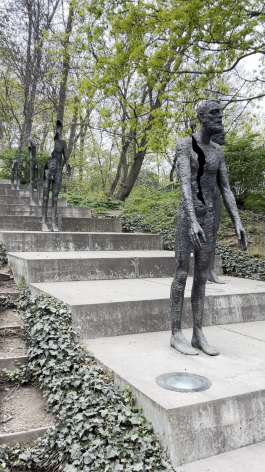} &
        \includegraphics[width=0.15\linewidth]{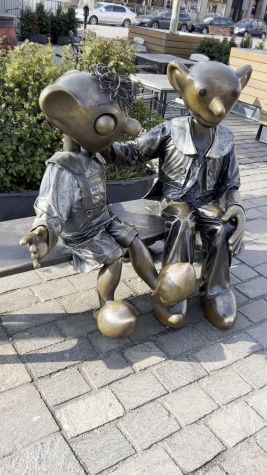} \\
        babies cerny & chair vltava & huge chair & memorial & spejbl hurvinek 
    \end{tabular}
    
    \vspace{1em}
    
    % Row 3 (1 image)
    \begin{subfigure}[b]{0.30\linewidth}
        \centering
        \includegraphics[width=\linewidth]{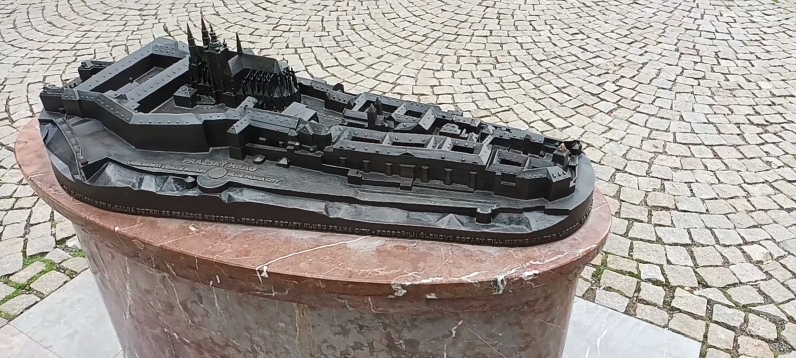}
        \caption*{prague castle}
    \end{subfigure}
    
    \caption{Overview of the 12 scenes included in the D2P-Statues subset of the D2P dataset.}
    \label{fig:statues_overview}
\end{figure*}

\begin{figure*}[t]
    \centering
    \begin{tabular}{ccccc}
        \includegraphics[width=0.15\linewidth]{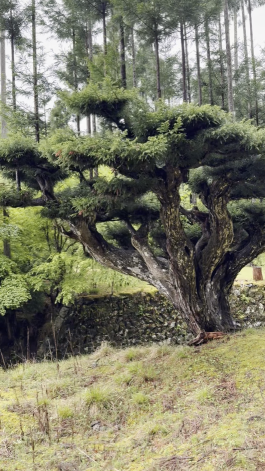} &
        \includegraphics[width=0.15\linewidth]{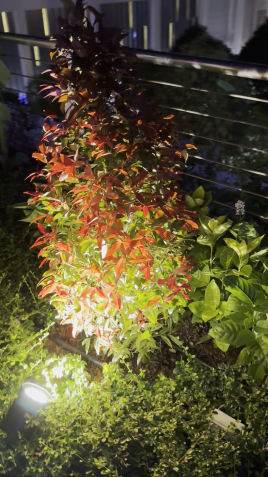} &
        \includegraphics[width=0.15\linewidth]{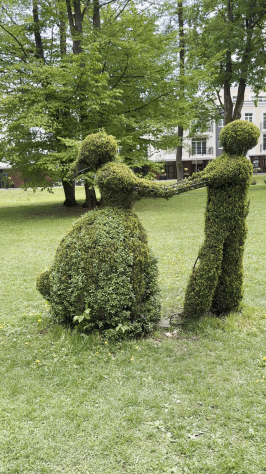} &
        \includegraphics[width=0.15\linewidth]{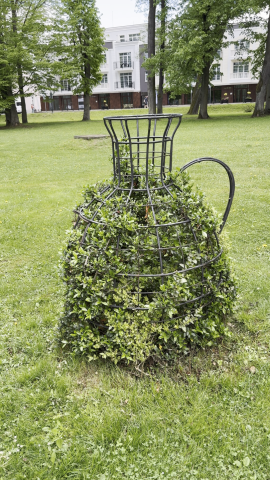} &
        \includegraphics[width=0.15\linewidth]{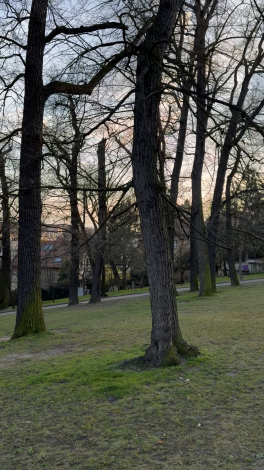} \\
        kyoto tree & \begin{tabular}{c} nadina \\ domestica \end{tabular} & \begin{tabular}{c} overgrown \\ couple \end{tabular} & overgrown pot & tree orbit \\
    \end{tabular}

    \vspace{1em}

    \begin{tabular}{ccccc}
        \includegraphics[width=0.15\linewidth]{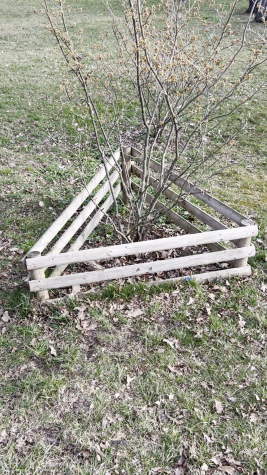} &
        \includegraphics[width=0.15\linewidth]{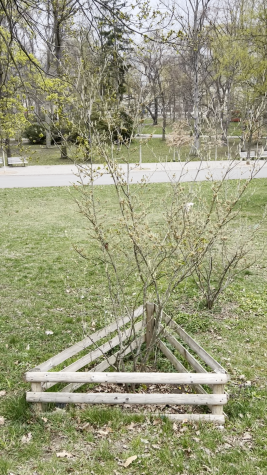} &
        \includegraphics[width=0.15\linewidth]{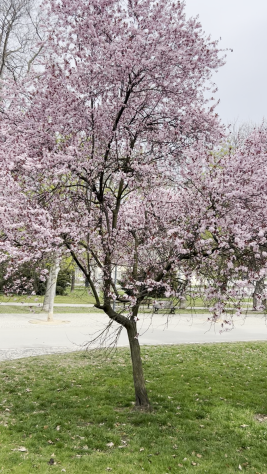} &
        \includegraphics[width=0.15\linewidth]{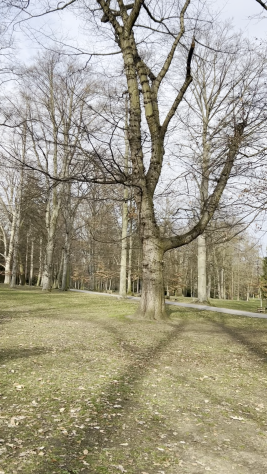} &
        \includegraphics[width=0.15\linewidth]{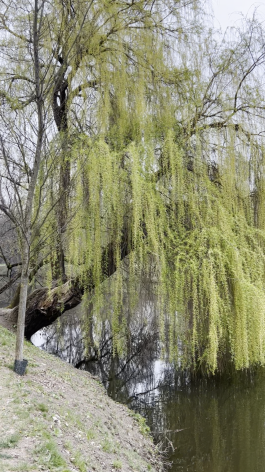} \\
        stromovka 2 & stromovka 2b & stromovka 5 & stromovka 3 & stromovka 7
    \end{tabular}
    
    \vspace{1em}

    \begin{tabular}{cc}   
    
    % Row 3 (1 image)
    \includegraphics[width=0.3\linewidth]{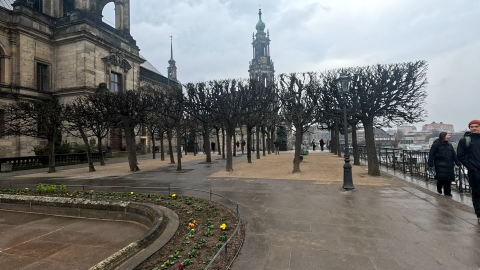} & \includegraphics[width=0.3\linewidth]{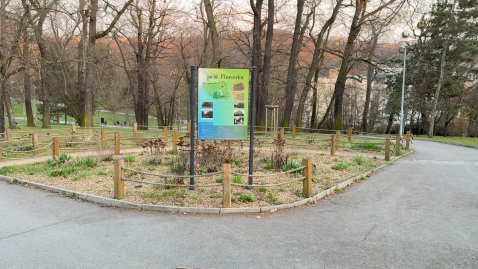} \\
    dresden & klamovka
    \end{tabular}
    
    \caption{Overview of the 12 scenes included in the D2P-Vegetation subset of the D2P dataset.}
    \label{fig:vegetation_overview}
\end{figure*}

The names of the scenes and representative sample images are shown in Fig.~\ref{fig:statues_overview} for the D2P-Statues subset and in Fig.~\ref{fig:vegetation_overview} for the D2P-Vegetation subset. For each scene in the D2P dataset, we sample 250 valid image pairs. The corresponding COLMAP reconstructions are visualized in Fig.~\ref{fig:d2p_colmap}.

\subsection{Pose Estimation}

\begin{table}[t]
    \centering
    \caption{The list of pose estimators used. Subscript $f$ denotes solvers for the uncalibrated case, $s$ denotes estimators that model shift. For scoring and LO S denotes the Sampson error, R the symmetric reprojection error and S+R their combination.}
    \begin{tabular}{cccccccc}    
    Estimator & Solver & Uses Depth & Affine-invariant & Uses $\M K$ & Scoring & LO+BA \\ \hline
    \baselinecalib & 5-point \cite{nister2004efficient} & \ding{55} & - & \checkmark & S & S\\
    \baselinesf & 6-point \cite{stewenius2008minimal} & \ding{55} & - & \ding{55} & S & S\\ \hline
    \calib & P3P~\cite{ding2023revisiting} & \checkmark & \ding{55} & \checkmark & S & S+R \\
    \calibshift & 3PT$_{suv}$~\cite{ding2025reposed} & \checkmark & \checkmark & \checkmark & S & S+R \\
    \mysf & 3PT$_{s00f}$~\cite{ding2025reposed} & \checkmark & \ding{55} & \ding{55} & S & S+R \\
    \sfshift & 4PT$_{suvf}$~\cite{ding2025reposed} & \checkmark & \checkmark & \ding{55} & S & S+R \\ \hline
    \calibro & P3P~\cite{ding2023revisiting} & \checkmark & \ding{55} & \checkmark & R & R \\
    \calibshiftro & 3PT$_{suv}$~\cite{ding2025reposed} & \checkmark & \checkmark & \checkmark & R & R \\
    \sfro & 3PT$_{s00f}$~\cite{ding2025reposed} & \checkmark & \ding{55} & \ding{55} & R & R \\
    \sfshiftro & 4PT$_{suvf}$~\cite{ding2025reposed} & \checkmark & \checkmark & \ding{55} & R & R \\ \hline
    \mdecalib & P3P~\cite{ding2023revisiting} & \checkmark & \ding{55} & \
    \multirow{4}{*}{%
  \begin{tabular}{@{}c@{}} $\M K$ \\ from \\ MDE \end{tabular}%
} & S & S+R \\
    \mdecalibshift & 3PT$_{suv}$~\cite{ding2025reposed} & \checkmark & \checkmark &  & S & S+R \\
    \mdecalibro & P3P~\cite{ding2023revisiting} & \checkmark & \ding{55} &  & R & R \\
    \mdecalibshiftro & 3PT$_{suv}$~\cite{ding2025reposed} & \checkmark & \checkmark &  & R & R \\ \hline
    \end{tabular}    
    \label{tab:estimators}
\end{table}

In addition to the estimators presented in the main paper. We include an additional baseline \baselinesf{} for the uncalibrated setting based on the PoseLib~\cite{PoseLib} implementation using 6-point solver~\cite{stewenius2008minimal}.
We also include the solver for unknown shared focal lenghts from RePoseD~\cite{ding2025reposed} denoted as \mysf. For the uncalibrated case we also consider an estimator using the calibrated RePoseD solver in combination with intrinsics estimated by MDEs where such output is provided, which we denote as \mdecalib. We do not combine the intrinsics-aware depth prediction with uncalibrated pose solvers, as this would represent an unrealistic setting in which calibration is available for depth estimation but not for pose recovery. In addition to the estimators which assume scale-invariant depths we also consider estimators which assume affine-invaraint depth estimates for each version. All of the estimators along with their properties are shown in Table~\ref{tab:estimators}.

All estimators are implemented within PoseLib~\cite{PoseLib} utilizing the LO-RANSAC framework~\cite{lebeda2012fixing}. For LO we use the truncated Sampson error (S) or truncated symmetric reprojection error loss. For estimators utilizing the hybrid error (S+R in Table~\ref{tab:estimators}) we use the truncated Cauchy loss for 100 iterations for the final BA as recommended on the official code repository of RePoseD~\cite{ding2025reposed}. For the remaining estimators we use the Cauchy loss.

\section{Additional Results}

\label{sec:more_results}

\subsection{Comparison with the Standard Benchmarks}

\begin{figure}
    \centering
\begin{center}
    Matches: LoMa~\cite{nordstrom2026loma}
\end{center}

\includegraphics[width=0.24\linewidth]{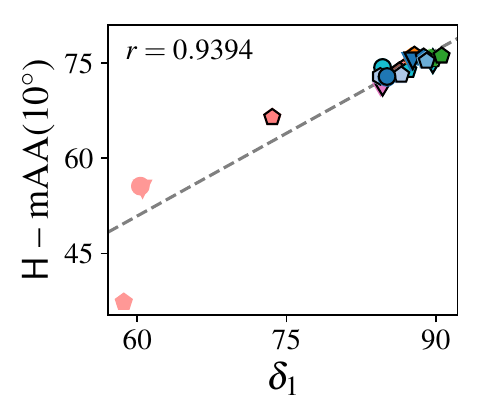} \hfill
\includegraphics[width=0.24\linewidth]{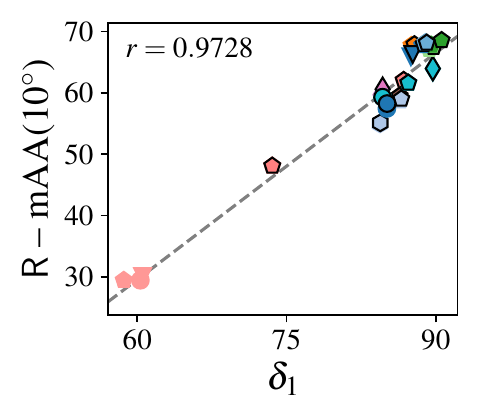} \hfill
\includegraphics[width=0.24\linewidth]{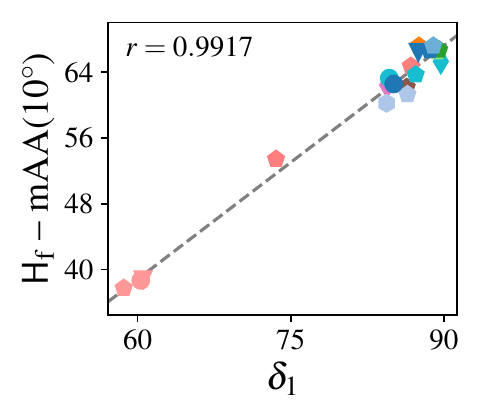} \hfill 
\includegraphics[width=0.24\linewidth]{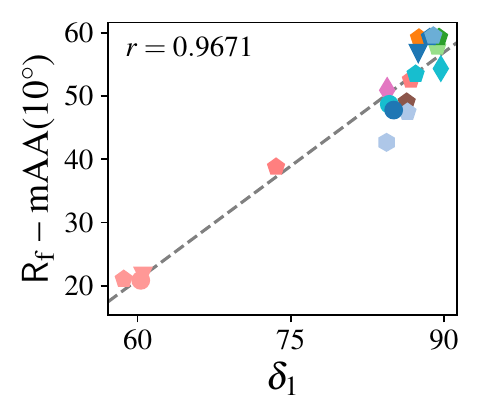}

\includegraphics[width=0.24\linewidth]{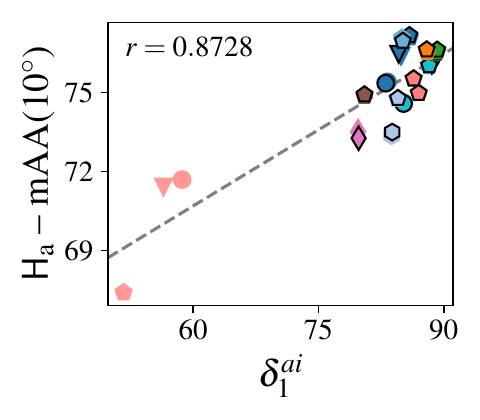} \hfill
\includegraphics[width=0.24\linewidth]{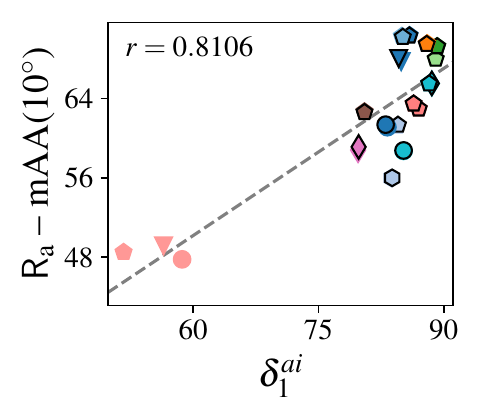} \hfill
\includegraphics[width=0.24\linewidth]{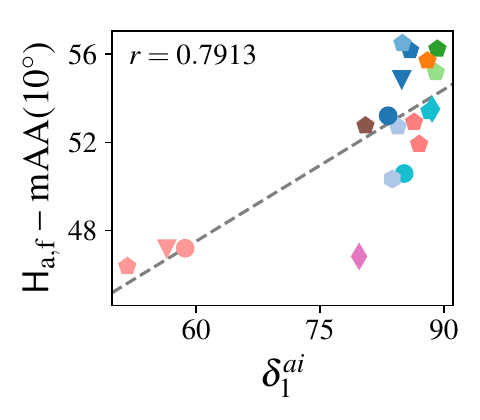} \hfill
\includegraphics[width=0.24\linewidth]{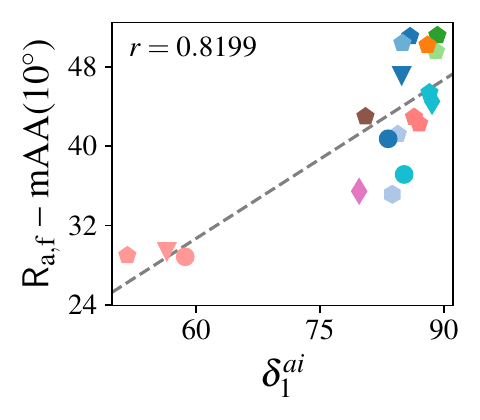} \hfill

\begin{center}
    Matches: SuperPoint~\cite{detone2018superpoint} + LightGlue~\cite{lindenberger2023lightglue}
\end{center}

\includegraphics[width=0.24\linewidth]{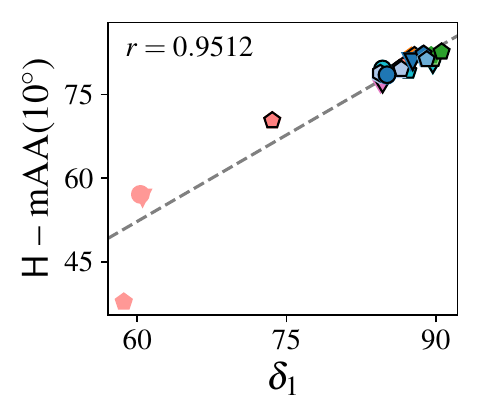} \hfill
\includegraphics[width=0.24\linewidth]{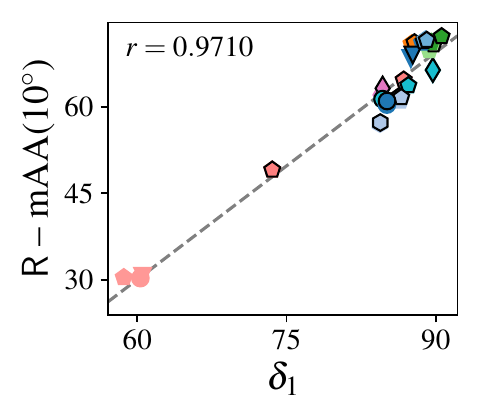} \hfill
\includegraphics[width=0.24\linewidth]{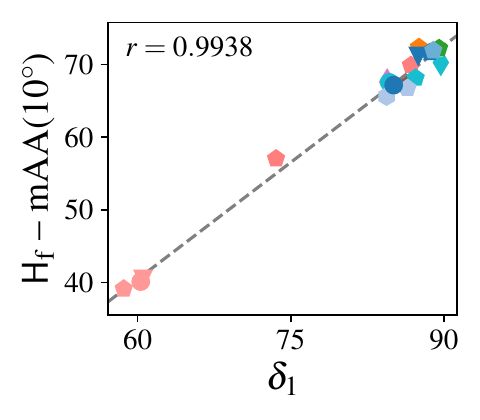} \hfill 
\includegraphics[width=0.24\linewidth]{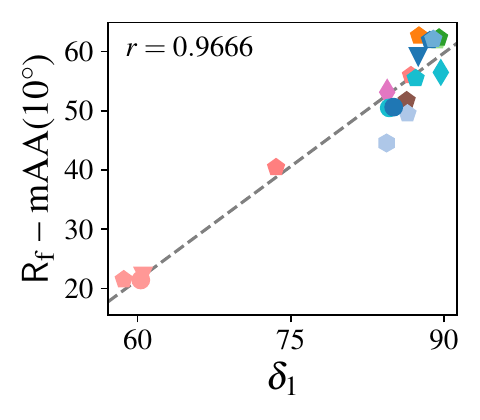}
\\

\includegraphics[width=0.24\linewidth]{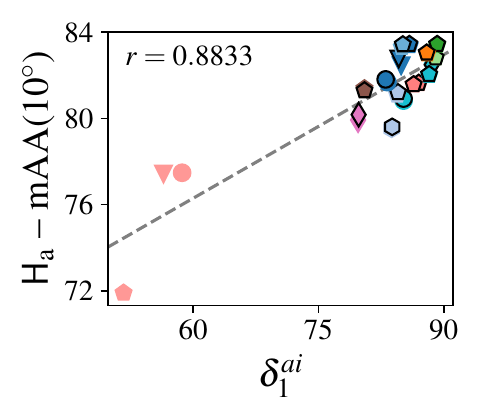} \hfill
\includegraphics[width=0.24\linewidth]{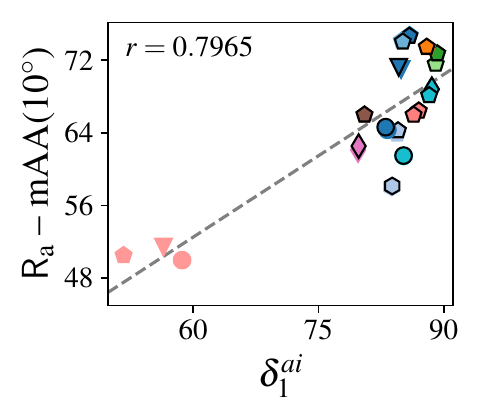} \hfill
\includegraphics[width=0.24\linewidth]{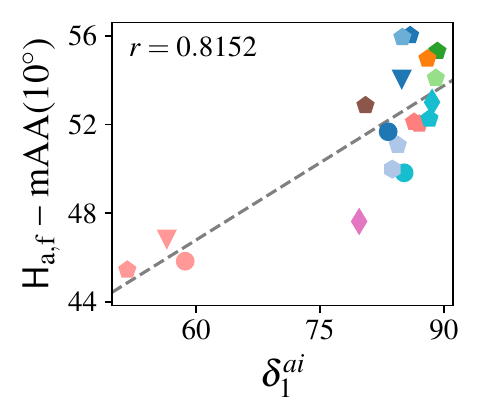} \hfill
\includegraphics[width=0.24\linewidth]{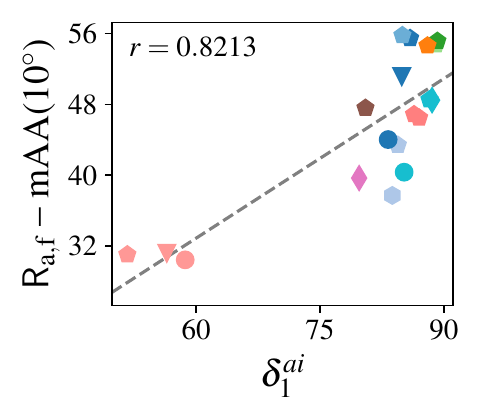} \hfill

\scriptsize
\begin{tabular}{*{6}{c}}        
    \includegraphics[width=12pt, valign=m]{figs/legend/UniDepth1.pdf} UniDepth1 
    & \includegraphics[width=12pt, valign=m]{figs/legend/UniDepth2.pdf} UniDepth2     
    & \includegraphics[width=12pt, valign=m]{figs/legend/UniK3D.pdf} UniK3D     
    & \includegraphics[width=12pt, valign=m]{figs/legend/MoGeV1.pdf} MoGeV1 
    & \includegraphics[width=12pt, valign=m]{figs/legend/MoGeV2.pdf} MoGeV2   
    & \includegraphics[width=12pt, valign=m]{figs/legend/Metric3DV2.pdf} Metric3DV2 
    
\end{tabular}
\vspace{-1ex}

\begin{tabular}{*{4}{c}}    
    \includegraphics[width=12pt, valign=m]{figs/legend/DAv3-Mono.pdf} DAv3-Mono 
    & \includegraphics[width=12pt, valign=m]{figs/legend/DAv3-Metric.pdf} DAv3-Metric     
    & \includegraphics[width=12pt, valign=m]{figs/legend/Pi3.pdf} Pi3 
    & \includegraphics[width=12pt, valign=m]{figs/legend/MapAnything.pdf} MapAnything
\end{tabular}
\vspace{-1ex}

\begin{tabular}{rccccc}
Backbone:
& \includegraphics[width=12pt, valign=m]{figs/legend/ViT-S.pdf} ViT-S
& \includegraphics[width=12pt, valign=m]{figs/legend/ViT-B.pdf} ViT-B
& \includegraphics[width=12pt, valign=m]{figs/legend/ViT-L.pdf} ViT-L
& \includegraphics[width=12pt, valign=m]{figs/legend/ViT-G.pdf} ViT-G
& \includegraphics[width=12pt, valign=m]{figs/legend/ConvNext.pdf} ConvNext
\end{tabular}

\begin{tabular}{rcc}
Uses Intrinsics at Inference: 
& \includegraphics[width=12pt, valign=m]{figs/legend/Calib.pdf} Yes
& \includegraphics[width=12pt, valign=m]{figs/legend/NoCalib.pdf} No
\end{tabular}

    \caption{The values of \dsi and \dssi compared agains \mAA for different estimators averaged over the standard benchmark datasets (\eth, \lamar, \sintel, \scannetpp). The plot includes all of the evaluated MDEs and point matches. The dashed line represents the linear fit of the data (the correlation coefficient $r$ is provided in the plots).}
    \label{fig:corr_plot_sm}
\end{figure}

To evaluate the experiments under the assumption that the depth is affine-invariant we include the metrics \dssi{} and \relssi{} which represent the same metrics, but under the affine-invariant assumption.

In Fig.~\ref{fig:corr_plot_sm} we show the same type of correlation plots as Fig.~\ref{fig:corr_plot} including matches obtained using~\cite{detone2018superpoint,lindenberger2023lightglue}. The results show that \dsi{} is still correlated to pose accuracy even when different matches are utilized. The experiments also show that the correlation is weaker when the depth is considered to be affine invariant. 

\begin{figure*}[!t]
  \centering
  \includegraphics[width=\linewidth]{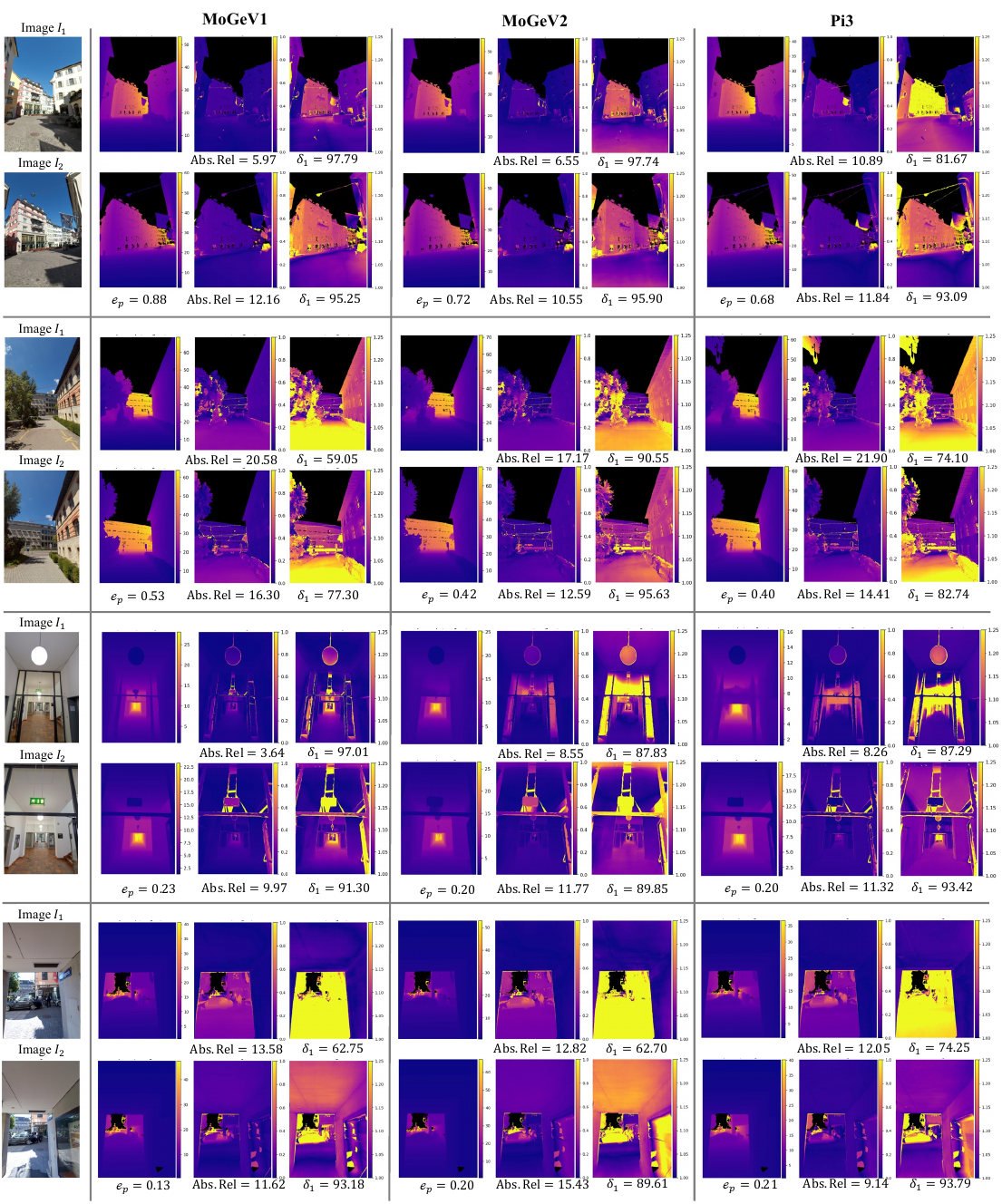}  
  \caption{Additional qualitative comparison of image pairs from the \lamar{} dataset. For each MDE, we show the predicted depth map together with error maps under the \relsi{} and \dsi{} metrics. For \dsi{}, all regions with a maximum ratio of 1.25 or larger are color-coded in yellow. $e_p$ denotes the relative pose error using our metric.}
  \label{fig:lamar_samples_appendix}
\end{figure*}

Additional qualitative comparisons of depth maps and error metrics are also shown in Fig.~\ref{fig:lamar_samples_appendix}.
As seen in the figure, high-error regions in depth prediction do not necessarily degrade pose accuracy because relative pose relies primarily on geometrically consistent correspondences rather than uniform depth accuracy across the entire image.
Large depth errors often occur in textureless areas, reflective surfaces, or distant background regions, which typically contribute few reliable feature matches and are therefore naturally down-weighted or rejected during geometric verification (RANSAC).
As a result, pose estimation is dominated by a subset of stable, well-textured regions where depth predictions preserve correct relative geometry, allowing accurate pose recovery even when substantial local depth errors are present elsewhere in the scene.

The extend results are also shown in Table~\ref{tab:sm_standard_calib_loma} for the calibrated case and Table~\ref{tab:sm_standard_uncal} for the uncalibrated case. And for individual scenes in Table~\ref{tab:sm_standard_per_scene}. The results for the affine-invariant case show that the estimated poses under this assumption can be more accurate. However, there is no significant change in the order of the methods with the exception of DAv3-Mono which is trained to produce affine-invariant depth and thus performs better in terms of \dssi{}, \relssi{} and also pose obtained using \calibshift{} and \calibshiftro{}. We also observe large \relsi{} errors for DepthPro and DepthAnythingV2 which are significantly improved when using \relssi instead, suggesting that the large errros are caused by the predicted depth not being scale-invariant, but affine-invariant.

\begin{table}[]
\caption{Results on the standard benchmark dataset (see Sec.~\ref{sec:standard_benchmark_info}) for the calibrated case using LoMa~\cite{nordstrom2026loma} matches.}
\label{tab:sm_standard_calib_loma}
\resizebox{\linewidth}{!}{
\begin{tabular}{lccccccccc}
\toprule
\multirow{2}{*}{MDE-Backbone} & MDE & \multirow{2}{*}{\dsi} & \multirow{2}{*}{\relsi} & \multirow{2}{*}{\dssi} & \multirow{2}{*}{\relssi} & \multicolumn{4}{c}{\mAA}\\  \cmidrule{7-10}
  & w/$\M K$ &   &   &   &   & \calib{} & \calibro{} & \calibshift{} & \calibshiftro{}\\ \midrule
MoGeV2-L & \checkmark & \rank{90.59}{1}{35} & \phantom{1}\phantom{1}\rank{9.45}{1}{35} & \rank{89.22}{2}{35} & \rank{20.57}{5}{35} & \rank{82.65}{1}{35} & \rank{72.22}{1}{35} & \rank{83.45}{1}{35} & \rank{72.75}{7}{35} \\
MoGeV1-L & \checkmark & \rank{89.78}{2}{35} & \phantom{1}\rank{10.47}{7}{35} & \rank{89.04}{4}{35} & \rank{21.41}{7}{35} & \rank{81.35}{7}{35} & \rank{70.79}{9}{35} & \rank{82.82}{10}{35} & \rank{71.60}{9}{35} \\
Metric3DV2-G & \checkmark & \rank{89.71}{3}{35} & \phantom{1}\rank{10.34}{4}{35} & \rank{88.58}{6}{35} & \rank{21.52}{9}{35} & \rank{80.97}{11}{35} & \rank{66.38}{13}{35} & \rank{82.50}{13}{35} & \rank{68.81}{13}{35} \\
Metric3DV2-G &  & \rank{89.71}{4}{35} & \phantom{1}\rank{10.34}{5}{35} & \rank{88.58}{5}{35} & \rank{21.52}{10}{35} & \rank{80.93}{12}{35} & \rank{66.38}{14}{35} & \rank{82.56}{12}{35} & \rank{68.80}{14}{35} \\
MoGeV2-L &  & \rank{89.54}{5}{35} & \phantom{1}\phantom{1}\rank{9.89}{2}{35} & \rank{89.22}{1}{35} & \rank{20.57}{4}{35} & \rank{81.96}{5}{35} & \rank{71.24}{5}{35} & \rank{83.42}{4}{35} & \rank{72.75}{7}{35} \\
MoGeV1-L &  & \rank{89.35}{6}{35} & \phantom{1}\rank{10.44}{6}{35} & \rank{89.04}{3}{35} & \rank{21.41}{8}{35} & \rank{81.29}{8}{35} & \rank{69.80}{10}{35} & \rank{82.82}{9}{35} & \rank{71.58}{10}{35} \\
UniK3D-L & \checkmark & \rank{89.09}{7}{35} & \phantom{1}\rank{14.37}{23}{35} & \rank{85.11}{19}{35} & \rank{22.99}{14}{35} & \rank{81.27}{10}{35} & \rank{71.59}{3}{35} & \rank{83.43}{3}{35} & \rank{74.05}{4}{35} \\
UniK3D-L &  & \rank{88.97}{8}{35} & \phantom{1}\rank{14.59}{24}{35} & \rank{84.99}{20}{35} & \rank{22.67}{13}{35} & \rank{81.28}{9}{35} & \rank{71.66}{2}{35} & \rank{83.31}{6}{35} & \rank{74.14}{3}{35} \\
UniDepth2-L & \checkmark & \rank{88.78}{9}{35} & \phantom{1}\rank{12.88}{16}{35} & \rank{85.89}{16}{35} & \rank{22.60}{12}{35} & \rank{82.25}{2}{35} & \rank{71.37}{4}{35} & \rank{83.45}{1}{35} & \rank{74.68}{2}{35} \\
UniDepth2-L &  & \rank{88.63}{10}{35} & \phantom{1}\rank{12.88}{15}{35} & \rank{85.92}{15}{35} & \rank{22.48}{11}{35} & \rank{82.12}{3}{35} & \rank{71.13}{7}{35} & \rank{83.39}{5}{35} & \rank{74.75}{1}{35} \\
Pi3-L & \checkmark & \rank{87.86}{11}{35} & \phantom{1}\rank{10.24}{3}{35} & \rank{87.97}{10}{35} & \rank{18.90}{2}{35} & \rank{82.01}{4}{35} & \rank{71.16}{6}{35} & \rank{83.05}{7}{35} & \rank{73.47}{5}{35} \\
UniDepth2-B & \checkmark & \rank{87.69}{12}{35} & \phantom{1}\rank{12.71}{14}{35} & \rank{84.61}{22}{35} & \rank{20.80}{6}{35} & \rank{80.83}{14}{35} & \rank{69.00}{11}{35} & \rank{82.76}{11}{35} & \rank{71.23}{11}{35} \\
Pi3-L &  & \rank{87.56}{13}{35} & \phantom{1}\rank{10.65}{8}{35} & \rank{88.02}{9}{35} & \rank{18.59}{1}{35} & \rank{81.89}{6}{35} & \rank{70.94}{8}{35} & \rank{83.01}{8}{35} & \rank{73.44}{6}{35} \\
UniDepth2-B &  & \rank{87.49}{14}{35} & \phantom{1}\rank{12.42}{13}{35} & \rank{84.91}{21}{35} & \rank{19.79}{3}{35} & \rank{80.91}{13}{35} & \rank{68.38}{12}{35} & \rank{82.45}{14}{35} & \rank{71.00}{12}{35} \\
Metric3DV2-L & \checkmark & \rank{87.24}{15}{35} & \phantom{1}\rank{11.13}{10}{35} & \rank{88.26}{7}{35} & \rank{23.31}{16}{35} & \rank{79.22}{23}{35} & \rank{63.71}{17}{35} & \rank{82.05}{15}{35} & \rank{68.13}{16}{35} \\
Metric3DV2-L &  & \rank{87.24}{16}{35} & \phantom{1}\rank{11.13}{9}{35} & \rank{88.26}{8}{35} & \rank{23.31}{15}{35} & \rank{79.22}{24}{35} & \rank{63.70}{18}{35} & \rank{82.01}{16}{35} & \rank{68.14}{15}{35} \\
DAv3-Metric-L &  & \rank{86.78}{17}{35} & \phantom{1}\rank{12.26}{11}{35} & \rank{87.01}{11}{35} & \rank{28.88}{25}{35} & \rank{80.17}{15}{35} & \rank{64.68}{16}{35} & \rank{81.63}{20}{35} & \rank{66.43}{18}{35} \\
DAv3-Metric-L & \checkmark & \rank{86.78}{18}{35} & \phantom{1}\rank{12.26}{12}{35} & \rank{87.01}{12}{35} & \rank{28.88}{26}{35} & \rank{80.17}{16}{35} & \rank{64.68}{15}{35} & \rank{81.64}{19}{35} & \rank{66.43}{17}{35} \\
UniDepth1-L & \checkmark & \rank{86.53}{19}{35} & \phantom{1}\rank{13.31}{19}{35} & \rank{84.52}{23}{35} & \rank{29.44}{28}{35} & \rank{79.53}{21}{35} & \rank{61.64}{23}{35} & \rank{81.21}{25}{35} & \rank{64.25}{25}{35} \\
UniDepth1-L &  & \rank{86.44}{20}{35} & \phantom{1}\rank{13.34}{20}{35} & \rank{84.43}{24}{35} & \rank{29.33}{27}{35} & \rank{79.31}{22}{35} & \rank{61.06}{26}{35} & \rank{81.10}{26}{35} & \rank{63.89}{26}{35} \\
DepthPro-L &  & \rank{86.36}{21}{35} & \rank{208.04}{34}{35} & \rank{80.51}{29}{35} & \rank{37.82}{32}{35} & \rank{79.66}{18}{35} & \rank{61.85}{22}{35} & \rank{81.39}{23}{35} & \rank{65.98}{20}{35} \\
DepthPro-L & \checkmark & \rank{86.36}{22}{35} & \rank{201.07}{33}{35} & \rank{80.51}{30}{35} & \rank{37.79}{31}{35} & \rank{79.68}{17}{35} & \rank{61.86}{21}{35} & \rank{81.30}{24}{35} & \rank{65.99}{19}{35} \\
UniDepth2-S & \checkmark & \rank{85.11}{23}{35} & \phantom{1}\rank{14.10}{22}{35} & \rank{83.05}{28}{35} & \rank{23.82}{18}{35} & \rank{78.53}{27}{35} & \rank{61.01}{27}{35} & \rank{81.81}{17}{35} & \rank{64.63}{23}{35} \\
UniDepth2-S &  & \rank{85.09}{24}{35} & \phantom{1}\rank{13.99}{21}{35} & \rank{83.26}{27}{35} & \rank{23.78}{17}{35} & \rank{78.43}{28}{35} & \rank{60.39}{28}{35} & \rank{81.69}{18}{35} & \rank{64.33}{24}{35} \\
MapAnything-G & \checkmark & \rank{84.66}{25}{35} & \phantom{1}\rank{15.87}{28}{35} & \rank{79.81}{31}{35} & \rank{26.18}{21}{35} & \rank{77.46}{30}{35} & \rank{63.20}{19}{35} & \rank{80.17}{29}{35} & \rank{62.54}{27}{35} \\
Metric3DV2-S &  & \rank{84.66}{26}{35} & \phantom{1}\rank{13.29}{18}{35} & \rank{85.20}{18}{35} & \rank{26.82}{24}{35} & \rank{79.54}{20}{35} & \rank{61.32}{25}{35} & \rank{80.83}{28}{35} & \rank{61.45}{30}{35} \\
Metric3DV2-S & \checkmark & \rank{84.65}{27}{35} & \phantom{1}\rank{13.29}{17}{35} & \rank{85.20}{17}{35} & \rank{26.82}{23}{35} & \rank{79.57}{19}{35} & \rank{61.34}{24}{35} & \rank{80.92}{27}{35} & \rank{61.50}{29}{35} \\
MapAnything-G &  & \rank{84.45}{28}{35} & \phantom{1}\rank{15.66}{27}{35} & \rank{79.74}{32}{35} & \rank{26.47}{22}{35} & \rank{77.53}{29}{35} & \rank{62.07}{20}{35} & \rank{79.93}{30}{35} & \rank{62.12}{28}{35} \\
UniDepth1-C & \checkmark & \rank{84.43}{29}{35} & \phantom{1}\rank{14.66}{26}{35} & \rank{83.83}{25}{35} & \rank{25.11}{19}{35} & \rank{78.80}{25}{35} & \rank{57.27}{29}{35} & \rank{79.61}{31}{35} & \rank{58.14}{31}{35} \\
UniDepth1-C &  & \rank{84.39}{30}{35} & \phantom{1}\rank{14.64}{25}{35} & \rank{83.76}{26}{35} & \rank{25.29}{20}{35} & \rank{78.55}{26}{35} & \rank{57.10}{30}{35} & \rank{79.52}{32}{35} & \rank{58.03}{32}{35} \\
DAv3-Mono-L &  & \rank{73.56}{31}{35} & \phantom{1}\rank{19.03}{29}{35} & \rank{86.40}{13}{35} & \rank{30.42}{29}{35} & \rank{70.33}{32}{35} & \rank{49.07}{31}{35} & \rank{81.57}{22}{35} & \rank{65.98}{21}{35} \\
DAv3-Mono-L & \checkmark & \rank{73.56}{32}{35} & \phantom{1}\rank{19.03}{30}{35} & \rank{86.40}{14}{35} & \rank{30.43}{30}{35} & \rank{70.34}{31}{35} & \rank{49.07}{32}{35} & \rank{81.58}{21}{35} & \rank{65.97}{22}{35} \\
DepthAnythingV2-B &  & \rank{60.53}{33}{35} & \rank{102.49}{32}{35} & \rank{56.44}{34}{35} & \rank{66.34}{34}{35} & \rank{56.48}{34}{35} & \rank{30.61}{33}{35} & \rank{77.42}{34}{35} & \rank{51.41}{33}{35} \\
DepthAnythingV2-S &  & \rank{60.31}{34}{35} & \phantom{1}\rank{72.70}{31}{35} & \rank{58.67}{33}{35} & \rank{60.30}{33}{35} & \rank{57.11}{33}{35} & \rank{30.28}{35}{35} & \rank{77.48}{33}{35} & \rank{49.97}{35}{35} \\
DepthAnythingV2-L &  & \rank{58.64}{35}{35} & \rank{275.18}{35}{35} & \rank{51.66}{35}{35} & \rank{72.81}{35}{35} & \rank{37.87}{35}{35} & \rank{30.40}{34}{35} & \rank{71.91}{35}{35} & \rank{50.50}{34}{35} \\
\midrule GT &  &   &   &   &   & \rank{90.70}{0}{35} & \rank{93.73}{0}{35} & \rank{89.57}{0}{35} & \rank{92.86}{0}{35} \\
\midrule No Depth + \baselinecalib{}&&&&&&\multicolumn{4}{c}{84.01}\\
\bottomrule
\end{tabular}
}
\end{table}

\begin{table}[]
    \centering
    \caption{Results on the standard benchmark dataset  (see Sec.~\ref{sec:standard_benchmark_info}) for the uncalibrated case.}
    \label{tab:sm_standard_uncal}
    
    Matches: LoMa~\cite{nordstrom2026loma}\\
    \vspace{2ex}
    
    \resizebox{\linewidth}{!}{
    \begin{tabular}{lcccccccccccc}
\toprule
\multirow{2}{*}{MDE-Backbone} & \multirow{2}{*}{\dsi} & \multirow{2}{*}{\relsi} & \multirow{2}{*}{\dssi} & \multirow{2}{*}{\relssi} & \multicolumn{8}{c}{\mAA}\\  \cmidrule{6-13}
  &   &   &   &   & \mysf{} & \sfro{} & \mdecalib{} & \mdecalibro{} & \sfshift{} & \sfshiftro{} & \mdecalibshift{} & \mdecalibshiftro{}\\ \midrule
Metric3DV2-G & \rank{89.71}{1}{19} & \phantom{1}\rank{10.34}{2}{19} & \rank{88.58}{3}{19} & \rank{21.52}{5}{19} & \rank{70.28}{7}{19} & \rank{56.49}{7}{19} &   &   & \rank{53.01}{7}{19} & \rank{48.49}{8}{19} &   &   \\
MoGeV2-L & \rank{89.54}{2}{19} & \phantom{1}\phantom{1}\rank{9.89}{1}{19} & \rank{89.22}{1}{19} & \rank{20.57}{3}{19} & \rank{72.23}{2}{19} & \rank{62.35}{2}{19} & \rank{56.14}{1}{19} & \rank{50.41}{1}{19} & \rank{55.30}{3}{19} & \rank{55.16}{3}{19} & \rank{57.90}{1}{19} & \rank{53.01}{1}{19} \\
MoGeV1-L & \rank{89.35}{3}{19} & \phantom{1}\rank{10.44}{3}{19} & \rank{89.04}{2}{19} & \rank{21.41}{4}{19} & \rank{72.08}{3}{19} & \rank{61.85}{5}{19} & \rank{54.15}{2}{19} & \rank{48.25}{2}{19} & \rank{54.09}{5}{19} & \rank{54.75}{4}{19} & \rank{55.28}{2}{19} & \rank{50.73}{2}{19} \\
UniK3D-L & \rank{88.97}{4}{19} & \phantom{1}\rank{14.59}{12}{19} & \rank{84.99}{10}{19} & \rank{22.67}{7}{19} & \rank{71.94}{4}{19} & \rank{62.12}{3}{19} &   &   & \rank{55.93}{2}{19} & \rank{55.81}{1}{19} &   &   \\
UniDepth2-L & \rank{88.63}{5}{19} & \phantom{1}\rank{12.88}{8}{19} & \rank{85.92}{8}{19} & \rank{22.48}{6}{19} & \rank{71.69}{5}{19} & \rank{61.92}{4}{19} & \rank{50.56}{4}{19} & \rank{46.01}{4}{19} & \rank{56.02}{1}{19} & \rank{55.48}{2}{19} & \rank{50.53}{4}{19} & \rank{48.37}{4}{19} \\
Pi3-L & \rank{87.56}{6}{19} & \phantom{1}\rank{10.65}{4}{19} & \rank{88.02}{5}{19} & \rank{18.59}{1}{19} & \rank{72.38}{1}{19} & \rank{62.68}{1}{19} & \rank{52.79}{3}{19} & \rank{46.56}{3}{19} & \rank{54.97}{4}{19} & \rank{54.64}{5}{19} & \rank{53.63}{3}{19} & \rank{49.50}{3}{19} \\
UniDepth2-B & \rank{87.49}{7}{19} & \phantom{1}\rank{12.42}{7}{19} & \rank{84.91}{11}{19} & \rank{19.79}{2}{19} & \rank{71.17}{6}{19} & \rank{59.17}{6}{19} & \rank{42.60}{6}{19} & \rank{38.52}{5}{19} & \rank{54.05}{6}{19} & \rank{51.16}{6}{19} & \rank{43.68}{5}{19} & \rank{41.10}{5}{19} \\
Metric3DV2-L & \rank{87.24}{8}{19} & \phantom{1}\rank{11.13}{5}{19} & \rank{88.26}{4}{19} & \rank{23.31}{8}{19} & \rank{68.15}{9}{19} & \rank{55.49}{9}{19} &   &   & \rank{52.26}{9}{19} & \rank{48.55}{7}{19} &   &   \\
DAv3-Metric-L & \rank{86.78}{9}{19} & \phantom{1}\rank{12.26}{6}{19} & \rank{87.01}{6}{19} & \rank{28.88}{13}{19} & \rank{69.91}{8}{19} & \rank{55.98}{8}{19} &   &   & \rank{52.03}{11}{19} & \rank{46.50}{11}{19} &   &   \\
UniDepth1-L & \rank{86.44}{10}{19} & \phantom{1}\rank{13.34}{10}{19} & \rank{84.43}{12}{19} & \rank{29.33}{14}{19} & \rank{66.75}{14}{19} & \rank{49.53}{14}{19} & \rank{20.12}{10}{19} & \rank{16.93}{10}{19} & \rank{51.07}{13}{19} & \rank{43.42}{13}{19} & \rank{20.31}{10}{19} & \rank{18.29}{9}{19} \\
DepthPro-L & \rank{86.36}{11}{19} & \rank{208.04}{18}{19} & \rank{80.51}{15}{19} & \rank{37.82}{16}{19} & \rank{67.10}{13}{19} & \rank{51.75}{11}{19} & \rank{27.80}{8}{19} & \rank{22.30}{8}{19} & \rank{52.86}{8}{19} & \rank{47.60}{9}{19} & \rank{29.19}{8}{19} & \rank{26.46}{8}{19} \\
UniDepth2-S & \rank{85.09}{12}{19} & \phantom{1}\rank{13.99}{11}{19} & \rank{83.26}{14}{19} & \rank{23.78}{9}{19} & \rank{67.16}{12}{19} & \rank{50.62}{12}{19} & \rank{34.50}{7}{19} & \rank{28.45}{7}{19} & \rank{51.67}{12}{19} & \rank{44.04}{12}{19} & \rank{35.45}{7}{19} & \rank{30.63}{7}{19} \\
Metric3DV2-S & \rank{84.66}{13}{19} & \phantom{1}\rank{13.29}{9}{19} & \rank{85.20}{9}{19} & \rank{26.82}{12}{19} & \rank{67.54}{11}{19} & \rank{50.50}{13}{19} &   &   & \rank{49.82}{15}{19} & \rank{40.35}{14}{19} &   &   \\
MapAnything-G & \rank{84.45}{14}{19} & \phantom{1}\rank{15.66}{14}{19} & \rank{79.74}{16}{19} & \rank{26.47}{11}{19} & \rank{67.61}{10}{19} & \rank{53.15}{10}{19} & \rank{43.65}{5}{19} & \rank{37.00}{6}{19} & \rank{47.63}{16}{19} & \rank{39.69}{15}{19} & \rank{42.99}{6}{19} & \rank{37.64}{6}{19} \\
UniDepth1-C & \rank{84.39}{15}{19} & \phantom{1}\rank{14.64}{13}{19} & \rank{83.76}{13}{19} & \rank{25.29}{10}{19} & \rank{65.61}{15}{19} & \rank{44.56}{15}{19} & \rank{20.89}{9}{19} & \rank{17.52}{9}{19} & \rank{49.99}{14}{19} & \rank{37.72}{16}{19} & \rank{20.55}{9}{19} & \rank{17.36}{10}{19} \\
DAv3-Mono-L & \rank{73.56}{16}{19} & \phantom{1}\rank{19.03}{15}{19} & \rank{86.40}{7}{19} & \rank{30.42}{15}{19} & \rank{57.05}{16}{19} & \rank{40.44}{16}{19} &   &   & \rank{52.11}{10}{19} & \rank{46.88}{10}{19} &   &   \\
DepthAnythingV2-B & \rank{60.53}{17}{19} & \rank{102.49}{17}{19} & \rank{56.44}{18}{19} & \rank{66.34}{18}{19} & \rank{40.46}{17}{19} & \rank{22.08}{17}{19} &   &   & \rank{46.84}{17}{19} & \rank{31.20}{17}{19} &   &   \\
DepthAnythingV2-S & \rank{60.31}{18}{19} & \phantom{1}\rank{72.70}{16}{19} & \rank{58.67}{17}{19} & \rank{60.30}{17}{19} & \rank{40.12}{18}{19} & \rank{21.42}{19}{19} &   &   & \rank{45.84}{18}{19} & \rank{30.45}{19}{19} &   &   \\
DepthAnythingV2-L & \rank{58.64}{19}{19} & \rank{275.18}{19}{19} & \rank{51.66}{19}{19} & \rank{72.81}{19}{19} & \rank{39.13}{19}{19} & \rank{21.50}{18}{19} &   &   & \rank{45.45}{19}{19} & \rank{31.05}{18}{19} &   &   \\
\midrule GT &   &   &   &   & \rank{87.16}{0}{19} & \rank{92.86}{0}{19} &   &   & \rank{70.18}{0}{19} & \rank{84.72}{0}{19} &   &   \\
\midrule No Depth + \baselinesf{}&&&&&\multicolumn{8}{c}{73.68}\\
\bottomrule
\end{tabular}}\\

\vspace{2ex}

Matches: SP~\cite{detone2018superpoint} + LG~\cite{lindenberger2023lightglue}\\

\vspace{2ex}

\resizebox{\linewidth}{!}{
\begin{tabular}{lcccccccccccc}
\toprule
\multirow{2}{*}{MDE-Backbone} & \multirow{2}{*}{\dsi} & \multirow{2}{*}{\relsi} & \multirow{2}{*}{\dssi} & \multirow{2}{*}{\relssi} & \multicolumn{8}{c}{\mAA}\\  \cmidrule{6-13}
  &   &   &   &   & \mysf{} & \sfro{} & \mdecalib{} & \mdecalibro{} & \sfshift{} & \sfshiftro{} & \mdecalibshift{} & \mdecalibshiftro{}\\ \midrule
Metric3DV2-G & \rank{89.71}{1}{19} & \phantom{1}\rank{10.34}{2}{19} & \rank{88.58}{3}{19} & \rank{21.52}{5}{19} & \rank{65.33}{7}{19} & \rank{54.33}{7}{19} &   &   & \rank{53.51}{7}{19} & \rank{44.51}{8}{19} &   &   \\
MoGeV2-L & \rank{89.54}{2}{19} & \phantom{1}\phantom{1}\rank{9.89}{1}{19} & \rank{89.22}{1}{19} & \rank{20.57}{3}{19} & \rank{66.74}{3}{19} & \rank{59.28}{2}{19} & \rank{52.72}{1}{19} & \rank{47.95}{1}{19} & \rank{56.24}{2}{19} & \rank{51.16}{1}{19} & \rank{54.20}{1}{19} & \rank{50.93}{1}{19} \\
MoGeV1-L & \rank{89.35}{3}{19} & \phantom{1}\rank{10.44}{3}{19} & \rank{89.04}{2}{19} & \rank{21.41}{4}{19} & \rank{66.52}{5}{19} & \rank{57.75}{5}{19} & \rank{51.30}{2}{19} & \rank{46.88}{2}{19} & \rank{55.17}{5}{19} & \rank{49.55}{5}{19} & \rank{52.38}{2}{19} & \rank{48.77}{2}{19} \\
UniK3D-L & \rank{88.97}{4}{19} & \phantom{1}\rank{14.59}{12}{19} & \rank{84.99}{10}{19} & \rank{22.67}{7}{19} & \rank{67.22}{1}{19} & \rank{59.51}{1}{19} &   &   & \rank{56.49}{1}{19} & \rank{50.38}{3}{19} &   &   \\
UniDepth2-L & \rank{88.63}{5}{19} & \phantom{1}\rank{12.88}{8}{19} & \rank{85.92}{8}{19} & \rank{22.48}{6}{19} & \rank{66.68}{4}{19} & \rank{59.28}{2}{19} & \rank{47.92}{4}{19} & \rank{44.66}{4}{19} & \rank{56.17}{3}{19} & \rank{51.05}{2}{19} & \rank{48.15}{4}{19} & \rank{46.56}{4}{19} \\
Pi3-L & \rank{87.56}{6}{19} & \phantom{1}\rank{10.65}{4}{19} & \rank{88.02}{5}{19} & \rank{18.59}{1}{19} & \rank{67.20}{2}{19} & \rank{59.19}{4}{19} & \rank{50.41}{3}{19} & \rank{44.99}{3}{19} & \rank{55.70}{4}{19} & \rank{50.19}{4}{19} & \rank{51.22}{3}{19} & \rank{47.79}{3}{19} \\
UniDepth2-B & \rank{87.49}{7}{19} & \phantom{1}\rank{12.42}{7}{19} & \rank{84.91}{11}{19} & \rank{19.79}{2}{19} & \rank{66.42}{6}{19} & \rank{56.76}{6}{19} & \rank{41.49}{6}{19} & \rank{38.46}{5}{19} & \rank{54.85}{6}{19} & \rank{47.11}{6}{19} & \rank{42.27}{5}{19} & \rank{40.44}{5}{19} \\
Metric3DV2-L & \rank{87.24}{8}{19} & \phantom{1}\rank{11.13}{5}{19} & \rank{88.26}{4}{19} & \rank{23.31}{8}{19} & \rank{63.71}{9}{19} & \rank{53.48}{8}{19} &   &   & \rank{53.41}{8}{19} & \rank{45.39}{7}{19} &   &   \\
DAv3-Metric-L & \rank{86.78}{9}{19} & \phantom{1}\rank{12.26}{6}{19} & \rank{87.01}{6}{19} & \rank{28.88}{13}{19} & \rank{64.76}{8}{19} & \rank{52.55}{9}{19} &   &   & \rank{51.92}{13}{19} & \rank{42.26}{11}{19} &   &   \\
UniDepth1-L & \rank{86.44}{10}{19} & \phantom{1}\rank{13.34}{10}{19} & \rank{84.43}{12}{19} & \rank{29.33}{14}{19} & \rank{61.33}{14}{19} & \rank{47.47}{14}{19} & \rank{19.49}{10}{19} & \rank{16.62}{10}{19} & \rank{52.71}{12}{19} & \rank{41.20}{12}{19} & \rank{19.89}{10}{19} & \rank{18.23}{9}{19} \\
DepthPro-L & \rank{86.36}{11}{19} & \rank{208.04}{18}{19} & \rank{80.51}{15}{19} & \rank{37.82}{16}{19} & \rank{62.19}{13}{19} & \rank{49.07}{11}{19} & \rank{27.08}{8}{19} & \rank{22.11}{8}{19} & \rank{52.76}{11}{19} & \rank{42.98}{9}{19} & \rank{28.10}{8}{19} & \rank{25.71}{8}{19} \\
UniDepth2-S & \rank{85.09}{12}{19} & \phantom{1}\rank{13.99}{11}{19} & \rank{83.26}{14}{19} & \rank{23.78}{9}{19} & \rank{62.58}{11}{19} & \rank{47.77}{13}{19} & \rank{33.86}{7}{19} & \rank{28.20}{7}{19} & \rank{53.21}{9}{19} & \rank{40.73}{13}{19} & \rank{34.48}{7}{19} & \rank{29.90}{7}{19} \\
Metric3DV2-S & \rank{84.66}{13}{19} & \phantom{1}\rank{13.29}{9}{19} & \rank{85.20}{9}{19} & \rank{26.82}{12}{19} & \rank{63.28}{10}{19} & \rank{48.68}{12}{19} &   &   & \rank{50.59}{14}{19} & \rank{37.13}{14}{19} &   &   \\
MapAnything-G & \rank{84.45}{14}{19} & \phantom{1}\rank{15.66}{14}{19} & \rank{79.74}{16}{19} & \rank{26.47}{11}{19} & \rank{62.44}{12}{19} & \rank{50.89}{10}{19} & \rank{41.62}{5}{19} & \rank{36.23}{6}{19} & \rank{46.82}{18}{19} & \rank{35.45}{15}{19} & \rank{41.10}{6}{19} & \rank{36.05}{6}{19} \\
UniDepth1-C & \rank{84.39}{15}{19} & \phantom{1}\rank{14.64}{13}{19} & \rank{83.76}{13}{19} & \rank{25.29}{10}{19} & \rank{60.23}{15}{19} & \rank{42.67}{15}{19} & \rank{20.28}{9}{19} & \rank{17.68}{9}{19} & \rank{50.34}{15}{19} & \rank{35.13}{16}{19} & \rank{20.32}{9}{19} & \rank{17.66}{10}{19} \\
DAv3-Mono-L & \rank{73.56}{16}{19} & \phantom{1}\rank{19.03}{15}{19} & \rank{86.40}{7}{19} & \rank{30.42}{15}{19} & \rank{53.45}{16}{19} & \rank{38.79}{16}{19} &   &   & \rank{52.92}{10}{19} & \rank{42.91}{10}{19} &   &   \\
DepthAnythingV2-B & \rank{60.53}{17}{19} & \rank{102.49}{17}{19} & \rank{56.44}{18}{19} & \rank{66.34}{18}{19} & \rank{38.76}{17}{19} & \rank{21.55}{17}{19} &   &   & \rank{47.17}{17}{19} & \rank{29.36}{17}{19} &   &   \\
DepthAnythingV2-S & \rank{60.31}{18}{19} & \phantom{1}\rank{72.70}{16}{19} & \rank{58.67}{17}{19} & \rank{60.30}{17}{19} & \rank{38.69}{18}{19} & \rank{20.86}{19}{19} &   &   & \rank{47.20}{16}{19} & \rank{28.84}{19}{19} &   &   \\
DepthAnythingV2-L & \rank{58.64}{19}{19} & \rank{275.18}{19}{19} & \rank{51.66}{19}{19} & \rank{72.81}{19}{19} & \rank{37.77}{19}{19} & \rank{21.06}{18}{19} &   &   & \rank{46.38}{19}{19} & \rank{28.98}{18}{19} &   &   \\
\midrule GT &   &   &   &   & \rank{81.36}{0}{19} & \rank{89.77}{0}{19} &   &   & \rank{69.02}{0}{19} & \rank{83.50}{0}{19} &   &   \\
\midrule No Depth + \baselinesf{}&&&&&\multicolumn{8}{c}{66.88}\\
\bottomrule
\end{tabular}
}
\end{table}

\begin{table}[]
    \centering
    \caption{Results on the standard benchmark dataset  (see Sec.~\ref{sec:standard_benchmark_info}) for individual scenes for the calibrated case using LoMa matches~\cite{nordstrom2026loma}. For \calib{} the \mAA metric is shown.}
    \label{tab:sm_standard_per_scene}
\resizebox{\linewidth}{!}{
\begin{tabular}{lccccccccc}
\toprule
  &   & \multicolumn{2}{c}{Eth3d} & \multicolumn{2}{c}{Scannetpp} & \multicolumn{2}{c}{Lamar} & \multicolumn{2}{c}{Sintel} \\ \cmidrule(lr){3-4} \cmidrule(lr){5-6} \cmidrule(lr){7-8} \cmidrule(lr){9-10}
MDE-Backbone & w/$\M K$ & \dsi & \calib{} & \dsi & \calib{} & \dsi & \calib{} & \dsi & \calib{} \\ \midrule
MoGeV2-L & \checkmark & \rank{98.54}{9}{35} & \rank{86.14}{4}{35} & \rank{97.12}{2}{35} & \rank{85.48}{1}{35} & \rank{90.53}{4}{35} & \rank{91.90}{7}{35} & \rank{76.17}{1}{35} & \rank{67.06}{3}{35} \\
MoGeV1-L & \checkmark & \rank{98.79}{5}{35} & \rank{85.99}{6}{35} & \rank{96.39}{15}{35} & \rank{83.55}{11}{35} & \rank{90.65}{3}{35} & \rank{91.98}{4}{35} & \rank{73.28}{4}{35} & \rank{63.87}{19}{35} \\
Metric3DV2-G & \checkmark & \rank{98.05}{15}{35} & \rank{85.00}{11}{35} & \rank{96.41}{14}{35} & \rank{80.84}{21}{35} & \rank{90.42}{5}{35} & \rank{90.81}{14}{35} & \rank{73.95}{2}{35} & \rank{67.25}{1}{35} \\
Metric3DV2-G &  & \rank{98.05}{14}{35} & \rank{84.98}{12}{35} & \rank{96.41}{13}{35} & \rank{80.82}{22}{35} & \rank{90.42}{6}{35} & \rank{90.72}{15}{35} & \rank{73.95}{3}{35} & \rank{67.19}{2}{35} \\
MoGeV2-L &  & \rank{98.59}{7}{35} & \rank{86.05}{5}{35} & \rank{97.16}{1}{35} & \rank{85.24}{4}{35} & \rank{90.70}{2}{35} & \rank{91.96}{5}{35} & \rank{71.69}{5}{35} & \rank{64.59}{10}{35} \\
MoGeV1-L &  & \rank{98.78}{6}{35} & \rank{86.35}{3}{35} & \rank{96.37}{16}{35} & \rank{83.02}{14}{35} & \rank{90.72}{1}{35} & \rank{91.99}{3}{35} & \rank{71.54}{6}{35} & \rank{63.81}{20}{35} \\
UniK3D-L & \checkmark & \rank{99.12}{1}{35} & \rank{85.34}{10}{35} & \rank{97.02}{4}{35} & \rank{85.41}{2}{35} & \rank{90.14}{7}{35} & \rank{91.93}{6}{35} & \rank{70.08}{7}{35} & \rank{62.41}{24}{35} \\
UniK3D-L &  & \rank{99.10}{3}{35} & \rank{85.51}{9}{35} & \rank{97.03}{3}{35} & \rank{85.31}{3}{35} & \rank{90.09}{8}{35} & \rank{91.76}{8}{35} & \rank{69.67}{8}{35} & \rank{62.53}{23}{35} \\
UniDepth2-L & \checkmark & \rank{99.10}{2}{35} & \rank{85.53}{8}{35} & \rank{96.99}{6}{35} & \rank{85.17}{5}{35} & \rank{89.93}{9}{35} & \rank{91.60}{9}{35} & \rank{69.12}{9}{35} & \rank{66.70}{4}{35} \\
UniDepth2-L &  & \rank{99.06}{4}{35} & \rank{85.68}{7}{35} & \rank{96.99}{5}{35} & \rank{85.09}{6}{35} & \rank{89.92}{10}{35} & \rank{91.32}{10}{35} & \rank{68.56}{10}{35} & \rank{66.39}{5}{35} \\
Pi3-L & \checkmark & \rank{98.53}{10}{35} & \rank{86.55}{2}{35} & \rank{96.73}{7}{35} & \rank{85.00}{7}{35} & \rank{89.85}{11}{35} & \rank{92.28}{1}{35} & \rank{66.31}{19}{35} & \rank{64.20}{16}{35} \\
UniDepth2-B & \checkmark & \rank{98.10}{13}{35} & \rank{84.45}{17}{35} & \rank{96.46}{10}{35} & \rank{83.77}{9}{35} & \rank{88.53}{16}{35} & \rank{90.87}{12}{35} & \rank{67.68}{11}{35} & \rank{64.24}{15}{35} \\
Pi3-L &  & \rank{98.55}{8}{35} & \rank{86.58}{1}{35} & \rank{96.69}{8}{35} & \rank{84.44}{8}{35} & \rank{89.73}{14}{35} & \rank{92.19}{2}{35} & \rank{65.29}{20}{35} & \rank{64.36}{13}{35} \\
UniDepth2-B &  & \rank{97.98}{16}{35} & \rank{84.87}{13}{35} & \rank{96.47}{9}{35} & \rank{83.73}{10}{35} & \rank{88.54}{15}{35} & \rank{90.44}{16}{35} & \rank{66.99}{18}{35} & \rank{64.59}{10}{35} \\
Metric3DV2-L & \checkmark & \rank{98.51}{11}{35} & \rank{84.30}{18}{35} & \rank{95.97}{17}{35} & \rank{81.65}{17}{35} & \rank{89.84}{13}{35} & \rank{90.89}{11}{35} & \rank{64.63}{21}{35} & \rank{60.04}{28}{35} \\
Metric3DV2-L &  & \rank{98.51}{12}{35} & \rank{84.25}{19}{35} & \rank{95.97}{18}{35} & \rank{81.66}{16}{35} & \rank{89.84}{12}{35} & \rank{90.84}{13}{35} & \rank{64.63}{22}{35} & \rank{60.12}{27}{35} \\
DAv3-Metric-L &  & \rank{97.96}{17}{35} & \rank{84.67}{15}{35} & \rank{96.45}{12}{35} & \rank{83.21}{12}{35} & \rank{85.61}{23}{35} & \rank{89.25}{27}{35} & \rank{67.09}{14}{35} & \rank{63.57}{21}{35} \\
DAv3-Metric-L & \checkmark & \rank{97.96}{18}{35} & \rank{84.65}{16}{35} & \rank{96.45}{11}{35} & \rank{83.21}{12}{35} & \rank{85.61}{24}{35} & \rank{89.24}{28}{35} & \rank{67.09}{15}{35} & \rank{63.57}{21}{35} \\
UniDepth1-L & \checkmark & \rank{96.76}{19}{35} & \rank{83.08}{25}{35} & \rank{95.54}{23}{35} & \rank{80.10}{27}{35} & \rank{86.63}{19}{35} & \rank{90.14}{17}{35} & \rank{67.21}{12}{35} & \rank{64.82}{9}{35} \\
UniDepth1-L &  & \rank{96.75}{20}{35} & \rank{82.82}{27}{35} & \rank{95.44}{24}{35} & \rank{79.59}{28}{35} & \rank{86.35}{21}{35} & \rank{89.89}{22}{35} & \rank{67.21}{13}{35} & \rank{64.94}{8}{35} \\
DepthPro-L &  & \rank{95.53}{25}{35} & \rank{83.66}{22}{35} & \rank{95.60}{22}{35} & \rank{80.54}{26}{35} & \rank{87.21}{18}{35} & \rank{90.06}{18}{35} & \rank{67.08}{16}{35} & \rank{64.38}{12}{35} \\
DepthPro-L & \checkmark & \rank{95.53}{26}{35} & \rank{83.70}{21}{35} & \rank{95.61}{21}{35} & \rank{80.59}{25}{35} & \rank{87.21}{17}{35} & \rank{90.06}{18}{35} & \rank{67.08}{17}{35} & \rank{64.36}{13}{35} \\
UniDepth2-S & \checkmark & \rank{95.70}{21}{35} & \rank{83.03}{26}{35} & \rank{95.66}{19}{35} & \rank{81.43}{20}{35} & \rank{84.98}{27}{35} & \rank{89.49}{23}{35} & \rank{64.09}{26}{35} & \rank{60.19}{26}{35} \\
UniDepth2-S &  & \rank{95.56}{24}{35} & \rank{82.54}{28}{35} & \rank{95.66}{20}{35} & \rank{81.52}{19}{35} & \rank{84.94}{28}{35} & \rank{89.31}{24}{35} & \rank{64.19}{25}{35} & \rank{60.35}{25}{35} \\
MapAnything-G & \checkmark & \rank{95.06}{29}{35} & \rank{84.70}{14}{35} & \rank{94.07}{28}{35} & \rank{81.55}{18}{35} & \rank{86.38}{20}{35} & \rank{89.08}{29}{35} & \rank{63.13}{27}{35} & \rank{54.52}{30}{35} \\
Metric3DV2-S &  & \rank{95.68}{23}{35} & \rank{83.51}{24}{35} & \rank{94.91}{26}{35} & \rank{80.72}{23}{35} & \rank{85.38}{25}{35} & \rank{89.99}{21}{35} & \rank{62.66}{28}{35} & \rank{63.95}{18}{35} \\
Metric3DV2-S & \checkmark & \rank{95.68}{22}{35} & \rank{83.57}{23}{35} & \rank{94.91}{25}{35} & \rank{80.71}{24}{35} & \rank{85.37}{26}{35} & \rank{90.01}{20}{35} & \rank{62.66}{29}{35} & \rank{63.98}{17}{35} \\
MapAnything-G &  & \rank{95.03}{30}{35} & \rank{84.22}{20}{35} & \rank{94.11}{27}{35} & \rank{81.76}{15}{35} & \rank{86.32}{22}{35} & \rank{88.61}{30}{35} & \rank{62.35}{30}{35} & \rank{55.55}{29}{35} \\
UniDepth1-C & \checkmark & \rank{95.11}{28}{35} & \rank{81.55}{30}{35} & \rank{93.52}{29}{35} & \rank{79.01}{29}{35} & \rank{84.47}{29}{35} & \rank{88.52}{31}{35} & \rank{64.60}{23}{35} & \rank{66.11}{6}{35} \\
UniDepth1-C &  & \rank{95.44}{27}{35} & \rank{81.72}{29}{35} & \rank{93.44}{30}{35} & \rank{78.71}{30}{35} & \rank{84.47}{30}{35} & \rank{87.72}{32}{35} & \rank{64.21}{24}{35} & \rank{66.03}{7}{35} \\
DAv3-Mono-L &  & \rank{87.56}{32}{35} & \rank{79.11}{32}{35} & \rank{79.22}{32}{35} & \rank{60.98}{31}{35} & \rank{74.85}{31}{35} & \rank{89.27}{26}{35} & \rank{52.63}{31}{35} & \rank{51.98}{31}{35} \\
DAv3-Mono-L & \checkmark & \rank{87.56}{31}{35} & \rank{79.12}{31}{35} & \rank{79.22}{31}{35} & \rank{60.98}{31}{35} & \rank{74.85}{32}{35} & \rank{89.28}{25}{35} & \rank{52.63}{32}{35} & \rank{51.98}{31}{35} \\
DepthAnythingV2-B &  & \rank{76.27}{33}{35} & \rank{67.58}{34}{35} & \rank{54.93}{34}{35} & \rank{29.49}{34}{35} & \rank{68.84}{33}{35} & \rank{80.89}{34}{35} & \rank{42.08}{34}{35} & \rank{47.96}{34}{35} \\
DepthAnythingV2-S &  & \rank{76.02}{34}{35} & \rank{68.67}{33}{35} & \rank{55.14}{33}{35} & \rank{30.41}{33}{35} & \rank{68.26}{34}{35} & \rank{81.19}{33}{35} & \rank{41.82}{35}{35} & \rank{48.18}{33}{35} \\
DepthAnythingV2-L &  & \rank{72.88}{35}{35} & \rank{34.44}{35}{35} & \rank{50.15}{35}{35} & \rank{10.65}{35}{35} & \rank{68.03}{35}{35} & \rank{68.72}{35}{35} & \rank{43.51}{33}{35} & \rank{37.68}{35}{35} \\
\midrule GT &  &   & \rank{92.32}{0}{35} &   & \rank{88.83}{0}{35} &   & \rank{96.07}{0}{35} &   & \rank{85.60}{0}{35} \\
\midrule No Depth + \baselinecalib{} &  &   & \multicolumn{1}{c}{85.90} &   & \multicolumn{1}{c}{82.72} &   & \multicolumn{1}{c}{92.72} &   & \multicolumn{1}{c}{74.70} \\
\bottomrule
\end{tabular}
}
\end{table}

\subsection{Extended Evaluation on the D2P Dataset}
Tables~\ref{tab:sm_d2p_loma} and \ref{tab:sm_d2p_splg} show extended results on the D2P dataset for both types of matches. The affine-invariant estimators achieve slightly better results for the calibrated case, but outside of few MDEs which produce affine-invariant depth (\eg DAv3-Mono) the order remains similar. For the uncalibrated estimators the order of methods is very similar to the equivalent calbrated estimators. However, for this case the affine-invariant estimators perform worse than the scale-invariant versions. 

\begin{table}[]
    \centering
    \caption{Results on the D2P dataset and its two subsets using LoMa matches~\cite{nordstrom2026loma} for the calibrated (top) and uncalibrated case (bottom).}   
    \resizebox{\linewidth}{!}{
        \begin{tabular}{c}
        \begin{tabular}{lccccccccccccc}
\toprule
  &   & \multicolumn{12}{c}{\mAA} \\ \cmidrule(lr){3-14}
  & MDE & \multicolumn{4}{c}{Mean} & \multicolumn{4}{c}{Statues} & \multicolumn{4}{c}{Vegetation} \\ \cmidrule(lr){3-6} \cmidrule(lr){7-10} \cmidrule(lr){11-14}
MDE-Backbone & w/$\M K$ & \calib{} & \calibro{} & \calibshift{} & \calibshiftro{} & \calib{} & \calibro{} & \calibshift{} & \calibshiftro{} & \calib{} & \calibro{} & \calibshift{} & \calibshiftro{}\\ \midrule
MoGeV1-L &  & \rank{78.62}{1}{35} & \rank{65.95}{1}{35} & \rank{81.01}{1}{35} & \rank{68.82}{1}{35} & \rank{84.97}{4}{35} & \rank{75.04}{3}{35} & \rank{87.15}{3}{35} & \rank{76.83}{1}{35} & \rank{80.14}{3}{35} & \rank{60.21}{3}{35} & \rank{80.78}{3}{35} & \rank{62.35}{3}{35} \\
Pi3-L & \checkmark & \rank{78.26}{2}{35} & \rank{64.45}{2}{35} & \rank{80.73}{4}{35} & \rank{66.99}{5}{35} & \rank{86.27}{2}{35} & \rank{75.28}{2}{35} & \rank{87.49}{1}{35} & \rank{74.56}{3}{35} & \rank{76.50}{18}{35} & \rank{51.85}{15}{35} & \rank{78.54}{12}{35} & \rank{55.33}{13}{35} \\
Pi3-L &  & \rank{77.72}{3}{35} & \rank{63.15}{3}{35} & \rank{80.58}{5}{35} & \rank{66.28}{6}{35} & \rank{85.65}{3}{35} & \rank{73.85}{4}{35} & \rank{87.28}{2}{35} & \rank{73.73}{6}{35} & \rank{76.78}{15}{35} & \rank{52.31}{14}{35} & \rank{78.33}{14}{35} & \rank{54.87}{14}{35} \\
UniK3D-L & \checkmark & \rank{76.81}{4}{35} & \rank{61.94}{5}{35} & \rank{80.48}{6}{35} & \rank{68.22}{4}{35} & \rank{82.72}{6}{35} & \rank{67.94}{7}{35} & \rank{85.19}{9}{35} & \rank{72.41}{8}{35} & \rank{79.95}{4}{35} & \rank{60.30}{2}{35} & \rank{80.61}{4}{35} & \rank{63.75}{2}{35} \\
MoGeV1-L & \checkmark & \rank{76.62}{5}{35} & \rank{62.23}{4}{35} & \rank{80.98}{2}{35} & \rank{68.75}{2}{35} & \rank{86.39}{1}{35} & \rank{76.18}{1}{35} & \rank{87.01}{4}{35} & \rank{76.79}{2}{35} & \rank{76.63}{17}{35} & \rank{53.38}{11}{35} & \rank{80.79}{2}{35} & \rank{62.20}{4}{35} \\
UniDepth2-L & \checkmark & \rank{76.47}{6}{35} & \rank{59.91}{6}{35} & \rank{79.85}{8}{35} & \rank{65.88}{10}{35} & \rank{81.80}{10}{35} & \rank{64.83}{11}{35} & \rank{85.01}{10}{35} & \rank{69.96}{12}{35} & \rank{79.57}{5}{35} & \rank{58.91}{5}{35} & \rank{80.32}{6}{35} & \rank{61.66}{5}{35} \\
UniDepth2-L &  & \rank{75.97}{7}{35} & \rank{58.70}{11}{35} & \rank{80.09}{7}{35} & \rank{66.02}{9}{35} & \rank{80.86}{16}{35} & \rank{62.06}{12}{35} & \rank{85.25}{8}{35} & \rank{70.15}{11}{35} & \rank{80.19}{2}{35} & \rank{59.11}{4}{35} & \rank{80.47}{5}{35} & \rank{61.38}{6}{35} \\
MapAnything-G & \checkmark & \rank{75.15}{8}{35} & \rank{58.79}{10}{35} & \rank{78.15}{17}{35} & \rank{63.11}{11}{35} & \rank{81.31}{13}{35} & \rank{67.89}{8}{35} & \rank{83.97}{22}{35} & \rank{70.20}{10}{35} & \rank{77.43}{10}{35} & \rank{54.06}{9}{35} & \rank{76.96}{21}{35} & \rank{55.42}{12}{35} \\
UniDepth1-L & \checkmark & \rank{75.07}{9}{35} & \rank{49.96}{21}{35} & \rank{76.98}{23}{35} & \rank{53.11}{25}{35} & \rank{80.93}{14}{35} & \rank{54.22}{23}{35} & \rank{83.53}{25}{35} & \rank{56.73}{25}{35} & \rank{77.35}{12}{35} & \rank{47.84}{19}{35} & \rank{76.47}{25}{35} & \rank{49.71}{26}{35} \\
MoGeV2-L &  & \rank{75.03}{10}{35} & \rank{59.72}{7}{35} & \rank{79.81}{9}{35} & \rank{66.03}{8}{35} & \rank{82.01}{8}{35} & \rank{68.81}{6}{35} & \rank{86.46}{5}{35} & \rank{73.98}{5}{35} & \rank{78.83}{9}{35} & \rank{57.87}{6}{35} & \rank{79.38}{9}{35} & \rank{59.89}{8}{35} \\
MoGeV2-L & \checkmark & \rank{75.02}{11}{35} & \rank{59.70}{8}{35} & \rank{79.76}{10}{35} & \rank{66.12}{7}{35} & \rank{83.66}{5}{35} & \rank{71.82}{5}{35} & \rank{86.44}{6}{35} & \rank{74.21}{4}{35} & \rank{72.93}{27}{35} & \rank{48.66}{18}{35} & \rank{79.44}{8}{35} & \rank{59.92}{7}{35} \\
DepthPro-L & \checkmark & \rank{74.93}{12}{35} & \rank{56.38}{14}{35} & \rank{78.52}{15}{35} & \rank{60.46}{18}{35} & \rank{76.94}{28}{35} & \rank{60.59}{13}{35} & \rank{84.16}{20}{35} & \rank{66.08}{15}{35} & \rank{79.01}{7}{35} & \rank{52.36}{12}{35} & \rank{78.49}{13}{35} & \rank{53.80}{20}{35} \\
UniDepth1-L &  & \rank{74.89}{13}{35} & \rank{48.27}{22}{35} & \rank{76.48}{26}{35} & \rank{51.31}{28}{35} & \rank{81.31}{12}{35} & \rank{54.29}{22}{35} & \rank{83.25}{26}{35} & \rank{56.34}{27}{35} & \rank{77.23}{14}{35} & \rank{47.28}{22}{35} & \rank{76.03}{26}{35} & \rank{49.04}{27}{35} \\
DepthPro-L &  & \rank{74.86}{14}{35} & \rank{56.39}{13}{35} & \rank{78.40}{16}{35} & \rank{60.50}{17}{35} & \rank{76.91}{29}{35} & \rank{60.57}{14}{35} & \rank{84.11}{21}{35} & \rank{65.94}{16}{35} & \rank{79.09}{6}{35} & \rank{52.33}{13}{35} & \rank{78.10}{16}{35} & \rank{53.92}{19}{35} \\
UniK3D-L &  & \rank{74.74}{15}{35} & \rank{59.11}{9}{35} & \rank{80.84}{3}{35} & \rank{68.61}{3}{35} & \rank{80.76}{17}{35} & \rank{64.87}{10}{35} & \rank{85.69}{7}{35} & \rank{73.55}{7}{35} & \rank{80.72}{1}{35} & \rank{61.05}{1}{35} & \rank{81.12}{1}{35} & \rank{64.14}{1}{35} \\
Metric3DV2-G & \checkmark & \rank{74.32}{16}{35} & \rank{50.31}{20}{35} & \rank{77.38}{21}{35} & \rank{54.31}{24}{35} & \rank{82.15}{7}{35} & \rank{57.96}{19}{35} & \rank{84.89}{12}{35} & \rank{62.07}{22}{35} & \rank{71.84}{29}{35} & \rank{40.80}{31}{35} & \rank{74.62}{31}{35} & \rank{44.41}{31}{35} \\
MapAnything-G &  & \rank{74.23}{17}{35} & \rank{57.03}{12}{35} & \rank{78.06}{18}{35} & \rank{62.56}{12}{35} & \rank{80.90}{15}{35} & \rank{66.42}{9}{35} & \rank{84.31}{17}{35} & \rank{70.47}{9}{35} & \rank{76.77}{16}{35} & \rank{54.13}{8}{35} & \rank{76.89}{22}{35} & \rank{55.62}{11}{35} \\
Metric3DV2-G &  & \rank{74.14}{18}{35} & \rank{50.35}{19}{35} & \rank{77.32}{22}{35} & \rank{54.40}{22}{35} & \rank{81.88}{9}{35} & \rank{57.87}{20}{35} & \rank{84.93}{11}{35} & \rank{62.15}{21}{35} & \rank{71.65}{30}{35} & \rank{40.92}{30}{35} & \rank{74.68}{30}{35} & \rank{44.59}{29}{35} \\
UniDepth2-B &  & \rank{74.05}{19}{35} & \rank{54.15}{16}{35} & \rank{78.77}{13}{35} & \rank{61.43}{15}{35} & \rank{80.02}{22}{35} & \rank{59.10}{16}{35} & \rank{84.22}{18}{35} & \rank{65.25}{17}{35} & \rank{78.94}{8}{35} & \rank{54.98}{7}{35} & \rank{79.25}{10}{35} & \rank{58.54}{9}{35} \\
UniDepth2-B & \checkmark & \rank{73.83}{20}{35} & \rank{54.23}{15}{35} & \rank{78.75}{14}{35} & \rank{61.11}{16}{35} & \rank{80.31}{20}{35} & \rank{59.78}{15}{35} & \rank{84.17}{19}{35} & \rank{65.00}{20}{35} & \rank{77.38}{11}{35} & \rank{53.85}{10}{35} & \rank{79.09}{11}{35} & \rank{57.81}{10}{35} \\
Metric3DV2-L &  & \rank{73.19}{21}{35} & \rank{47.18}{23}{35} & \rank{76.19}{27}{35} & \rank{52.52}{27}{35} & \rank{78.90}{26}{35} & \rank{53.24}{25}{35} & \rank{82.22}{29}{35} & \rank{56.28}{28}{35} & \rank{70.55}{32}{35} & \rank{36.12}{34}{35} & \rank{72.44}{34}{35} & \rank{42.59}{35}{35} \\
Metric3DV2-L & \checkmark & \rank{73.13}{22}{35} & \rank{47.15}{24}{35} & \rank{75.96}{29}{35} & \rank{52.57}{26}{35} & \rank{78.97}{25}{35} & \rank{53.27}{24}{35} & \rank{82.03}{30}{35} & \rank{56.41}{26}{35} & \rank{70.59}{31}{35} & \rank{36.01}{35}{35} & \rank{71.71}{35}{35} & \rank{42.59}{34}{35} \\
DAv3-Metric-L &  & \rank{72.74}{23}{35} & \rank{50.90}{17}{35} & \rank{77.61}{20}{35} & \rank{57.55}{19}{35} & \rank{79.22}{23}{35} & \rank{58.54}{17}{35} & \rank{84.85}{14}{35} & \rank{65.09}{18}{35} & \rank{75.66}{21}{35} & \rank{47.82}{20}{35} & \rank{76.77}{24}{35} & \rank{53.11}{22}{35} \\
DAv3-Metric-L & \checkmark & \rank{72.73}{24}{35} & \rank{50.90}{18}{35} & \rank{77.62}{19}{35} & \rank{57.52}{20}{35} & \rank{79.22}{23}{35} & \rank{58.54}{18}{35} & \rank{84.85}{13}{35} & \rank{65.09}{19}{35} & \rank{75.64}{22}{35} & \rank{47.80}{21}{35} & \rank{76.82}{23}{35} & \rank{53.04}{24}{35} \\
Metric3DV2-S & \checkmark & \rank{72.28}{25}{35} & \rank{44.96}{26}{35} & \rank{75.78}{30}{35} & \rank{50.32}{30}{35} & \rank{80.36}{19}{35} & \rank{50.72}{27}{35} & \rank{82.86}{28}{35} & \rank{55.75}{30}{35} & \rank{72.83}{28}{35} & \rank{40.77}{32}{35} & \rank{74.13}{32}{35} & \rank{43.50}{33}{35} \\
Metric3DV2-S &  & \rank{72.26}{26}{35} & \rank{44.87}{27}{35} & \rank{76.13}{28}{35} & \rank{50.35}{29}{35} & \rank{80.28}{21}{35} & \rank{50.66}{28}{35} & \rank{83.13}{27}{35} & \rank{55.75}{29}{35} & \rank{73.06}{26}{35} & \rank{40.73}{33}{35} & \rank{74.83}{29}{35} & \rank{43.54}{32}{35} \\
UniDepth1-C &  & \rank{71.48}{27}{35} & \rank{40.53}{30}{35} & \rank{74.41}{32}{35} & \rank{44.86}{35}{35} & \rank{76.90}{30}{35} & \rank{44.09}{29}{35} & \rank{80.33}{34}{35} & \rank{47.30}{35}{35} & \rank{76.23}{19}{35} & \rank{41.28}{25}{35} & \rank{75.86}{28}{35} & \rank{44.48}{30}{35} \\
UniDepth1-C & \checkmark & \rank{71.03}{28}{35} & \rank{40.86}{29}{35} & \rank{74.81}{31}{35} & \rank{46.54}{33}{35} & \rank{77.47}{27}{35} & \rank{43.89}{30}{35} & \rank{80.61}{33}{35} & \rank{47.92}{34}{35} & \rank{76.02}{20}{35} & \rank{41.97}{24}{35} & \rank{76.03}{27}{35} & \rank{45.05}{28}{35} \\
UniDepth2-S & \checkmark & \rank{69.81}{29}{35} & \rank{45.95}{25}{35} & \rank{76.65}{25}{35} & \rank{54.35}{23}{35} & \rank{81.37}{11}{35} & \rank{55.62}{21}{35} & \rank{83.65}{24}{35} & \rank{61.33}{24}{35} & \rank{75.28}{23}{35} & \rank{49.29}{17}{35} & \rank{77.69}{20}{35} & \rank{53.11}{23}{35} \\
UniDepth2-S &  & \rank{69.60}{30}{35} & \rank{44.82}{28}{35} & \rank{76.85}{24}{35} & \rank{55.02}{21}{35} & \rank{80.42}{18}{35} & \rank{53.12}{26}{35} & \rank{83.67}{23}{35} & \rank{61.84}{23}{35} & \rank{77.24}{13}{35} & \rank{50.49}{16}{35} & \rank{78.10}{17}{35} & \rank{53.31}{21}{35} \\
DAv3-Mono-L & \checkmark & \rank{50.22}{31}{35} & \rank{26.06}{33}{35} & \rank{78.93}{11}{35} & \rank{62.49}{13}{35} & \rank{54.34}{33}{35} & \rank{25.82}{33}{35} & \rank{84.58}{15}{35} & \rank{69.01}{13}{35} & \rank{75.02}{24}{35} & \rank{41.23}{27}{35} & \rank{78.19}{15}{35} & \rank{54.12}{17}{35} \\
DAv3-Mono-L &  & \rank{50.21}{32}{35} & \rank{26.06}{32}{35} & \rank{78.88}{12}{35} & \rank{62.48}{14}{35} & \rank{54.35}{32}{35} & \rank{25.82}{34}{35} & \rank{84.57}{16}{35} & \rank{69.00}{14}{35} & \rank{74.98}{25}{35} & \rank{41.24}{26}{35} & \rank{78.05}{18}{35} & \rank{54.08}{18}{35} \\
DepthAnythingV2-S &  & \rank{47.68}{33}{35} & \rank{24.73}{34}{35} & \rank{73.35}{34}{35} & \rank{45.85}{34}{35} & \rank{55.93}{31}{35} & \rank{26.06}{32}{35} & \rank{81.35}{31}{35} & \rank{51.49}{33}{35} & \rank{69.44}{33}{35} & \rank{41.18}{28}{35} & \rank{78.00}{19}{35} & \rank{52.05}{25}{35} \\
DepthAnythingV2-B &  & \rank{46.22}{34}{35} & \rank{24.05}{35}{35} & \rank{74.14}{33}{35} & \rank{48.23}{32}{35} & \rank{51.53}{34}{35} & \rank{23.98}{35}{35} & \rank{81.05}{32}{35} & \rank{52.71}{32}{35} & \rank{69.28}{34}{35} & \rank{41.10}{29}{35} & \rank{79.48}{7}{35} & \rank{54.41}{16}{35} \\
DepthAnythingV2-L &  & \rank{20.68}{35}{35} & \rank{26.27}{31}{35} & \rank{61.13}{35}{35} & \rank{50.09}{31}{35} & \rank{15.34}{35}{35} & \rank{26.33}{31}{35} & \rank{66.49}{35}{35} & \rank{54.56}{31}{35} & \rank{44.29}{35}{35} & \rank{43.48}{23}{35} & \rank{72.92}{33}{35} & \rank{54.73}{15}{35} \\
\midrule GT &  & \rank{93.39}{0}{35} & \rank{97.65}{0}{35} & \rank{92.52}{0}{35} & \rank{97.20}{0}{35} & \rank{95.80}{0}{35} & \rank{98.87}{0}{35} & \rank{95.19}{0}{35} & \rank{98.67}{0}{35} & \rank{94.53}{0}{35} & \rank{97.69}{0}{35} & \rank{93.87}{0}{35} & \rank{97.29}{0}{35} \\
\midrule No Depth + \baselinecalib{} &  & \multicolumn{4}{c}{82.34} & \multicolumn{4}{c}{88.78} & \multicolumn{4}{c}{86.31} \\
\bottomrule
\end{tabular} 
        \\    
        ~
        \\
        \begin{tabular}{lcccccccccccc}
\toprule
  & \multicolumn{12}{c}{\mAA} \\ \cmidrule(lr){2-13}
  & \multicolumn{4}{c}{Mean} & \multicolumn{4}{c}{Statues} & \multicolumn{4}{c}{Vegetation} \\ \cmidrule(lr){2-5} \cmidrule(lr){6-9} \cmidrule(lr){10-13}
MDE-Backbone & \mysf{} & \sfro{} & \sfshift{} & \sfshiftro{} & \mysf{} & \sfro{} & \sfshift{} & \sfshiftro{} & \mysf{} & \sfro{} & \sfshift{} & \sfshiftro{}\\ \midrule
MoGeV1-L & \rank{73.27}{1}{19} & \rank{61.81}{1}{19} & \rank{46.33}{2}{19} & \rank{42.90}{2}{19} & \rank{80.37}{1}{19} & \rank{72.69}{1}{19} & \rank{51.17}{1}{19} & \rank{48.76}{1}{19} & \rank{70.80}{3}{19} & \rank{51.40}{3}{19} & \rank{47.99}{4}{19} & \rank{35.63}{3}{19} \\
Pi3-L & \rank{71.39}{2}{19} & \rank{59.30}{2}{19} & \rank{45.28}{4}{19} & \rank{40.98}{5}{19} & \rank{79.88}{2}{19} & \rank{70.96}{2}{19} & \rank{49.46}{4}{19} & \rank{44.50}{4}{19} & \rank{66.39}{8}{19} & \rank{45.71}{7}{19} & \rank{45.15}{12}{19} & \rank{30.47}{8}{19} \\
UniK3D-L & \rank{70.40}{3}{19} & \rank{58.55}{3}{19} & \rank{46.62}{1}{19} & \rank{44.31}{1}{19} & \rank{75.24}{4}{19} & \rank{65.52}{4}{19} & \rank{49.85}{2}{19} & \rank{46.71}{3}{19} & \rank{72.25}{1}{19} & \rank{53.99}{1}{19} & \rank{50.18}{1}{19} & \rank{40.07}{1}{19} \\
UniDepth2-L & \rank{70.37}{4}{19} & \rank{56.65}{5}{19} & \rank{46.15}{3}{19} & \rank{42.54}{3}{19} & \rank{75.18}{5}{19} & \rank{62.37}{6}{19} & \rank{49.13}{5}{19} & \rank{44.25}{5}{19} & \rank{71.21}{2}{19} & \rank{51.67}{2}{19} & \rank{50.15}{2}{19} & \rank{38.20}{2}{19} \\
MoGeV2-L & \rank{69.36}{5}{19} & \rank{58.09}{4}{19} & \rank{45.25}{5}{19} & \rank{41.51}{4}{19} & \rank{76.61}{3}{19} & \rank{68.95}{3}{19} & \rank{49.47}{3}{19} & \rank{46.88}{2}{19} & \rank{68.41}{5}{19} & \rank{49.60}{4}{19} & \rank{47.72}{5}{19} & \rank{34.72}{5}{19} \\
DepthPro-L & \rank{67.88}{6}{19} & \rank{51.95}{8}{19} & \rank{44.09}{7}{19} & \rank{35.62}{8}{19} & \rank{72.55}{10}{19} & \rank{59.53}{7}{19} & \rank{47.97}{8}{19} & \rank{37.56}{9}{19} & \rank{66.83}{7}{19} & \rank{42.70}{9}{19} & \rank{46.03}{9}{19} & \rank{28.89}{13}{19} \\
MapAnything-G & \rank{67.73}{7}{19} & \rank{53.92}{6}{19} & \rank{42.23}{11}{19} & \rank{35.55}{9}{19} & \rank{74.55}{6}{19} & \rank{63.68}{5}{19} & \rank{46.88}{11}{19} & \rank{39.07}{8}{19} & \rank{66.94}{6}{19} & \rank{46.60}{6}{19} & \rank{44.28}{15}{19} & \rank{30.17}{11}{19} \\
UniDepth2-B & \rank{67.40}{8}{19} & \rank{52.05}{7}{19} & \rank{45.06}{6}{19} & \rank{38.32}{6}{19} & \rank{73.24}{8}{19} & \rank{58.54}{8}{19} & \rank{48.61}{6}{19} & \rank{39.08}{7}{19} & \rank{69.13}{4}{19} & \rank{47.08}{5}{19} & \rank{49.41}{3}{19} & \rank{35.23}{4}{19} \\
Metric3DV2-G & \rank{65.54}{9}{19} & \rank{46.22}{10}{19} & \rank{42.20}{12}{19} & \rank{29.68}{12}{19} & \rank{73.75}{7}{19} & \rank{55.23}{10}{19} & \rank{45.86}{12}{19} & \rank{32.69}{12}{19} & \rank{58.96}{16}{19} & \rank{34.62}{16}{19} & \rank{43.65}{17}{19} & \rank{23.08}{16}{19} \\
DAv3-Metric-L & \rank{65.44}{10}{19} & \rank{47.79}{9}{19} & \rank{42.83}{9}{19} & \rank{32.77}{10}{19} & \rank{72.85}{9}{19} & \rank{57.36}{9}{19} & \rank{47.56}{9}{19} & \rank{37.46}{10}{19} & \rank{64.81}{10}{19} & \rank{41.83}{10}{19} & \rank{44.63}{13}{19} & \rank{29.28}{12}{19} \\
UniDepth1-L & \rank{64.95}{11}{19} & \rank{43.14}{13}{19} & \rank{41.12}{13}{19} & \rank{27.02}{15}{19} & \rank{70.86}{12}{19} & \rank{50.74}{13}{19} & \rank{45.12}{13}{19} & \rank{28.48}{14}{19} & \rank{64.66}{11}{19} & \rank{38.07}{11}{19} & \rank{44.62}{14}{19} & \rank{26.35}{15}{19} \\
Metric3DV2-L & \rank{63.60}{12}{19} & \rank{44.64}{11}{19} & \rank{41.10}{14}{19} & \rank{28.61}{13}{19} & \rank{70.74}{13}{19} & \rank{51.81}{12}{19} & \rank{44.93}{14}{19} & \rank{30.55}{13}{19} & \rank{54.73}{19}{19} & \rank{31.56}{19}{19} & \rank{40.94}{19}{19} & \rank{22.35}{18}{19} \\
Metric3DV2-S & \rank{62.51}{13}{19} & \rank{40.96}{14}{19} & \rank{39.82}{15}{19} & \rank{25.33}{17}{19} & \rank{69.12}{14}{19} & \rank{48.09}{14}{19} & \rank{43.26}{17}{19} & \rank{27.06}{17}{19} & \rank{59.69}{15}{19} & \rank{32.80}{18}{19} & \rank{41.63}{18}{19} & \rank{20.85}{19}{19} \\
UniDepth2-S & \rank{60.98}{14}{19} & \rank{43.61}{12}{19} & \rank{42.47}{10}{19} & \rank{31.83}{11}{19} & \rank{71.89}{11}{19} & \rank{52.80}{11}{19} & \rank{47.22}{10}{19} & \rank{34.44}{11}{19} & \rank{66.10}{9}{19} & \rank{42.91}{8}{19} & \rank{46.97}{6}{19} & \rank{30.31}{10}{19} \\
UniDepth1-C & \rank{60.46}{15}{19} & \rank{35.50}{15}{19} & \rank{39.28}{17}{19} & \rank{22.98}{19}{19} & \rank{65.12}{15}{19} & \rank{40.01}{15}{19} & \rank{42.86}{18}{19} & \rank{23.02}{19}{19} & \rank{62.00}{13}{19} & \rank{33.25}{17}{19} & \rank{43.92}{16}{19} & \rank{22.74}{17}{19} \\
DAv3-Mono-L & \rank{44.61}{16}{19} & \rank{29.68}{17}{19} & \rank{44.06}{8}{19} & \rank{37.33}{7}{19} & \rank{45.74}{18}{19} & \rank{30.40}{19}{19} & \rank{48.06}{7}{19} & \rank{41.44}{6}{19} & \rank{63.25}{12}{19} & \rank{35.72}{14}{19} & \rank{45.31}{10}{19} & \rank{30.73}{7}{19} \\
DepthAnythingV2-L & \rank{43.91}{17}{19} & \rank{30.92}{16}{19} & \rank{38.60}{18}{19} & \rank{27.44}{14}{19} & \rank{46.22}{16}{19} & \rank{32.80}{16}{19} & \rank{41.08}{19}{19} & \rank{27.83}{15}{19} & \rank{60.00}{14}{19} & \rank{37.88}{12}{19} & \rank{46.07}{8}{19} & \rank{31.05}{6}{19} \\
DepthAnythingV2-S & \rank{41.35}{18}{19} & \rank{28.00}{19}{19} & \rank{38.45}{19}{19} & \rank{24.68}{18}{19} & \rank{45.99}{17}{19} & \rank{31.49}{17}{19} & \rank{43.45}{16}{19} & \rank{25.85}{18}{19} & \rank{57.22}{18}{19} & \rank{35.72}{15}{19} & \rank{45.26}{11}{19} & \rank{27.79}{14}{19} \\
DepthAnythingV2-B & \rank{41.15}{19}{19} & \rank{28.17}{18}{19} & \rank{39.58}{16}{19} & \rank{26.50}{16}{19} & \rank{44.11}{19}{19} & \rank{30.80}{18}{19} & \rank{44.25}{15}{19} & \rank{27.59}{16}{19} & \rank{57.76}{17}{19} & \rank{36.27}{13}{19} & \rank{46.81}{7}{19} & \rank{30.41}{9}{19} \\
\midrule GT & \rank{91.56}{0}{19} & \rank{96.98}{0}{19} & \rank{63.19}{0}{19} & \rank{82.75}{0}{19} & \rank{95.03}{0}{19} & \rank{98.42}{0}{19} & \rank{64.32}{0}{19} & \rank{83.57}{0}{19} & \rank{92.05}{0}{19} & \rank{96.69}{0}{19} & \rank{72.40}{0}{19} & \rank{86.43}{0}{19} \\
\midrule No Depth + \baselinesf{} & \multicolumn{4}{c}{67.92} & \multicolumn{4}{c}{76.22} & \multicolumn{4}{c}{76.09} \\
\bottomrule
\end{tabular}
        \end{tabular}
    }
    
    \label{tab:sm_d2p_loma}
\end{table}

\begin{table}[]
    \centering
    \caption{Results on the D2P dataset and its two subsets using SP~\cite{detone2018superpoint}+LG~\cite{lindenberger2023lightglue} matches for the calibrated (top) and uncalibrated case.}   
    \resizebox{\linewidth}{!}{
        \begin{tabular}{c}
        \begin{tabular}{lccccccccccccc}
\toprule
  &   & \multicolumn{12}{c}{\mAA} \\ \cmidrule(lr){3-14}
  & MDE & \multicolumn{4}{c}{Mean} & \multicolumn{4}{c}{Statues} & \multicolumn{4}{c}{Vegetation} \\ \cmidrule(lr){3-6} \cmidrule(lr){7-10} \cmidrule(lr){11-14}
MDE-Backbone & w/$\M K$ & \calib{} & \calibro{} & \calibshift{} & \calibshiftro{} & \calib{} & \calibro{} & \calibshift{} & \calibshiftro{} & \calib{} & \calibro{} & \calibshift{} & \calibshiftro{}\\ \midrule
MoGeV1-L &  & \rank{67.75}{1}{35} & \rank{57.19}{1}{35} & \rank{68.67}{2}{35} & \rank{59.27}{3}{35} & \rank{75.24}{4}{35} & \rank{66.13}{3}{35} & \rank{77.37}{1}{35} & \rank{69.69}{1}{35} & \rank{70.34}{1}{35} & \rank{53.37}{1}{35} & \rank{68.56}{2}{35} & \rank{52.49}{4}{35} \\
Pi3-L & \checkmark & \rank{67.19}{2}{35} & \rank{54.25}{3}{35} & \rank{68.29}{5}{35} & \rank{57.06}{7}{35} & \rank{76.16}{2}{35} & \rank{66.47}{2}{35} & \rank{77.36}{2}{35} & \rank{67.90}{3}{35} & \rank{66.06}{20}{35} & \rank{40.59}{17}{35} & \rank{66.23}{17}{35} & \rank{44.30}{17}{35} \\
Pi3-L &  & \rank{66.69}{3}{35} & \rank{53.38}{5}{35} & \rank{68.05}{7}{35} & \rank{56.52}{8}{35} & \rank{75.41}{3}{35} & \rank{64.84}{4}{35} & \rank{77.22}{4}{35} & \rank{66.98}{6}{35} & \rank{66.43}{17}{35} & \rank{41.22}{16}{35} & \rank{65.63}{20}{35} & \rank{44.21}{18}{35} \\
UniK3D-L & \checkmark & \rank{66.32}{4}{35} & \rank{54.66}{2}{35} & \rank{68.51}{4}{35} & \rank{59.10}{4}{35} & \rank{73.64}{6}{35} & \rank{62.03}{6}{35} & \rank{76.16}{7}{35} & \rank{66.16}{8}{35} & \rank{69.51}{2}{35} & \rank{52.25}{3}{35} & \rank{68.32}{3}{35} & \rank{53.38}{2}{35} \\
UniDepth2-L & \checkmark & \rank{66.20}{5}{35} & \rank{53.32}{6}{35} & \rank{67.83}{8}{35} & \rank{57.13}{6}{35} & \rank{73.27}{7}{35} & \rank{59.48}{10}{35} & \rank{75.41}{10}{35} & \rank{64.14}{9}{35} & \rank{68.28}{5}{35} & \rank{50.74}{5}{35} & \rank{68.08}{5}{35} & \rank{51.41}{6}{35} \\
UniDepth2-L &  & \rank{65.95}{6}{35} & \rank{52.57}{7}{35} & \rank{68.06}{6}{35} & \rank{57.33}{5}{35} & \rank{72.81}{8}{35} & \rank{57.55}{12}{35} & \rank{75.67}{9}{35} & \rank{63.90}{10}{35} & \rank{68.53}{4}{35} & \rank{50.95}{4}{35} & \rank{68.04}{6}{35} & \rank{51.92}{5}{35} \\
MoGeV1-L & \checkmark & \rank{65.63}{7}{35} & \rank{53.75}{4}{35} & \rank{68.61}{3}{35} & \rank{59.30}{2}{35} & \rank{76.50}{1}{35} & \rank{67.87}{1}{35} & \rank{77.24}{3}{35} & \rank{69.68}{2}{35} & \rank{67.53}{10}{35} & \rank{46.80}{9}{35} & \rank{68.28}{4}{35} & \rank{52.59}{3}{35} \\
MoGeV2-L & \checkmark & \rank{65.24}{8}{35} & \rank{50.73}{11}{35} & \rank{67.47}{9}{35} & \rank{55.81}{9}{35} & \rank{74.72}{5}{35} & \rank{64.49}{5}{35} & \rank{76.87}{5}{35} & \rank{67.22}{4}{35} & \rank{65.00}{24}{35} & \rank{39.25}{20}{35} & \rank{66.38}{14}{35} & \rank{47.02}{11}{35} \\
MapAnything-G & \checkmark & \rank{65.10}{9}{35} & \rank{51.86}{9}{35} & \rank{66.50}{17}{35} & \rank{55.09}{11}{35} & \rank{72.26}{9}{35} & \rank{59.88}{8}{35} & \rank{74.95}{15}{35} & \rank{63.05}{14}{35} & \rank{66.43}{16}{35} & \rank{47.40}{7}{35} & \rank{65.24}{21}{35} & \rank{48.07}{10}{35} \\
DepthPro-L & \checkmark & \rank{64.70}{10}{35} & \rank{48.23}{13}{35} & \rank{66.70}{13}{35} & \rank{52.11}{17}{35} & \rank{68.46}{29}{35} & \rank{54.34}{13}{35} & \rank{74.99}{13}{35} & \rank{61.79}{16}{35} & \rank{68.01}{7}{35} & \rank{42.07}{15}{35} & \rank{66.71}{11}{35} & \rank{42.94}{19}{35} \\
DepthPro-L &  & \rank{64.66}{11}{35} & \rank{48.21}{14}{35} & \rank{66.65}{14}{35} & \rank{52.00}{18}{35} & \rank{68.39}{30}{35} & \rank{54.19}{14}{35} & \rank{74.88}{16}{35} & \rank{61.82}{15}{35} & \rank{67.94}{8}{35} & \rank{42.12}{14}{35} & \rank{66.64}{12}{35} & \rank{42.60}{21}{35} \\
MapAnything-G &  & \rank{64.41}{12}{35} & \rank{51.03}{10}{35} & \rank{66.15}{18}{35} & \rank{54.73}{12}{35} & \rank{71.65}{14}{35} & \rank{59.12}{11}{35} & \rank{74.96}{14}{35} & \rank{63.46}{11}{35} & \rank{66.75}{14}{35} & \rank{48.27}{6}{35} & \rank{64.81}{28}{35} & \rank{48.40}{9}{35} \\
UniK3D-L &  & \rank{64.37}{13}{35} & \rank{52.19}{8}{35} & \rank{68.68}{1}{35} & \rank{59.40}{1}{35} & \rank{72.05}{11}{35} & \rank{59.59}{9}{35} & \rank{76.12}{8}{35} & \rank{66.79}{7}{35} & \rank{69.09}{3}{35} & \rank{52.41}{2}{35} & \rank{69.02}{1}{35} & \rank{54.23}{1}{35} \\
MoGeV2-L &  & \rank{64.12}{14}{35} & \rank{50.14}{12}{35} & \rank{67.45}{10}{35} & \rank{55.77}{10}{35} & \rank{72.21}{10}{35} & \rank{60.49}{7}{35} & \rank{76.77}{6}{35} & \rank{67.15}{5}{35} & \rank{67.06}{12}{35} & \rank{44.28}{12}{35} & \rank{66.21}{18}{35} & \rank{46.95}{12}{35} \\
UniDepth2-B &  & \rank{63.76}{15}{35} & \rank{47.24}{16}{35} & \rank{66.62}{15}{35} & \rank{53.66}{13}{35} & \rank{70.54}{20}{35} & \rank{51.93}{16}{35} & \rank{74.15}{17}{35} & \rank{58.66}{17}{35} & \rank{68.28}{6}{35} & \rank{46.91}{8}{35} & \rank{66.88}{10}{35} & \rank{49.53}{7}{35} \\
UniDepth2-B & \checkmark & \rank{63.66}{16}{35} & \rank{47.37}{15}{35} & \rank{66.61}{16}{35} & \rank{53.08}{16}{35} & \rank{70.86}{17}{35} & \rank{52.68}{15}{35} & \rank{74.09}{18}{35} & \rank{58.20}{18}{35} & \rank{67.24}{11}{35} & \rank{45.76}{10}{35} & \rank{66.89}{9}{35} & \rank{48.50}{8}{35} \\
UniDepth1-L & \checkmark & \rank{63.65}{17}{35} & \rank{42.62}{17}{35} & \rank{64.89}{23}{35} & \rank{45.26}{25}{35} & \rank{70.96}{16}{35} & \rank{47.34}{22}{35} & \rank{72.82}{27}{35} & \rank{51.88}{25}{35} & \rank{66.29}{19}{35} & \rank{40.25}{18}{35} & \rank{65.21}{22}{35} & \rank{40.67}{24}{35} \\
UniDepth1-L &  & \rank{63.60}{18}{35} & \rank{41.23}{22}{35} & \rank{64.62}{27}{35} & \rank{43.55}{28}{35} & \rank{71.49}{15}{35} & \rank{47.54}{21}{35} & \rank{73.09}{25}{35} & \rank{50.97}{26}{35} & \rank{66.45}{15}{35} & \rank{39.84}{19}{35} & \rank{64.98}{25}{35} & \rank{40.27}{27}{35} \\
Metric3DV2-S & \checkmark & \rank{63.09}{19}{35} & \rank{38.03}{27}{35} & \rank{64.81}{25}{35} & \rank{42.03}{30}{35} & \rank{70.55}{19}{35} & \rank{43.33}{27}{35} & \rank{73.66}{22}{35} & \rank{49.14}{30}{35} & \rank{64.87}{25}{35} & \rank{32.52}{31}{35} & \rank{63.62}{32}{35} & \rank{33.60}{34}{35} \\
Metric3DV2-S &  & \rank{63.07}{20}{35} & \rank{37.97}{28}{35} & \rank{64.88}{24}{35} & \rank{41.98}{31}{35} & \rank{70.52}{21}{35} & \rank{43.22}{28}{35} & \rank{73.73}{21}{35} & \rank{49.01}{31}{35} & \rank{64.67}{26}{35} & \rank{32.44}{32}{35} & \rank{63.72}{31}{35} & \rank{33.50}{35}{35} \\
Metric3DV2-G & \checkmark & \rank{62.99}{21}{35} & \rank{42.60}{18}{35} & \rank{65.00}{22}{35} & \rank{45.53}{23}{35} & \rank{72.04}{12}{35} & \rank{51.54}{17}{35} & \rank{73.28}{24}{35} & \rank{54.79}{21}{35} & \rank{61.99}{31}{35} & \rank{32.83}{29}{35} & \rank{63.41}{33}{35} & \rank{34.09}{32}{35} \\
Metric3DV2-G &  & \rank{62.98}{22}{35} & \rank{42.57}{19}{35} & \rank{65.24}{21}{35} & \rank{45.51}{24}{35} & \rank{72.02}{13}{35} & \rank{51.45}{18}{35} & \rank{73.58}{23}{35} & \rank{54.76}{22}{35} & \rank{61.84}{32}{35} & \rank{32.77}{30}{35} & \rank{63.84}{30}{35} & \rank{34.02}{33}{35} \\
Metric3DV2-L &  & \rank{62.68}{23}{35} & \rank{40.97}{24}{35} & \rank{64.24}{29}{35} & \rank{44.58}{26}{35} & \rank{69.06}{25}{35} & \rank{45.78}{26}{35} & \rank{72.59}{29}{35} & \rank{50.71}{27}{35} & \rank{62.21}{29}{35} & \rank{31.99}{34}{35} & \rank{62.00}{34}{35} & \rank{34.98}{30}{35} \\
Metric3DV2-L & \checkmark & \rank{62.62}{24}{35} & \rank{41.00}{23}{35} & \rank{63.99}{30}{35} & \rank{44.44}{27}{35} & \rank{69.03}{26}{35} & \rank{45.83}{25}{35} & \rank{72.52}{30}{35} & \rank{50.55}{28}{35} & \rank{62.00}{30}{35} & \rank{32.04}{33}{35} & \rank{61.46}{35}{35} & \rank{34.69}{31}{35} \\
UniDepth1-C &  & \rank{62.27}{25}{35} & \rank{35.73}{30}{35} & \rank{63.79}{32}{35} & \rank{39.89}{34}{35} & \rank{68.49}{28}{35} & \rank{38.19}{30}{35} & \rank{71.41}{34}{35} & \rank{44.51}{35}{35} & \rank{66.82}{13}{35} & \rank{37.27}{24}{35} & \rank{65.20}{23}{35} & \rank{38.71}{28}{35} \\
UniDepth1-C & \checkmark & \rank{61.68}{26}{35} & \rank{36.07}{29}{35} & \rank{63.97}{31}{35} & \rank{40.92}{33}{35} & \rank{68.73}{27}{35} & \rank{38.47}{29}{35} & \rank{71.65}{31}{35} & \rank{45.20}{34}{35} & \rank{66.40}{18}{35} & \rank{37.30}{23}{35} & \rank{64.84}{27}{35} & \rank{38.61}{29}{35} \\
DAv3-Metric-L &  & \rank{61.52}{27}{35} & \rank{41.88}{20}{35} & \rank{65.52}{19}{35} & \rank{47.80}{20}{35} & \rank{69.24}{24}{35} & \rank{50.75}{19}{35} & \rank{73.96}{20}{35} & \rank{57.83}{19}{35} & \rank{65.52}{22}{35} & \rank{38.61}{21}{35} & \rank{65.12}{24}{35} & \rank{40.59}{26}{35} \\
DAv3-Metric-L & \checkmark & \rank{61.50}{28}{35} & \rank{41.86}{21}{35} & \rank{65.47}{20}{35} & \rank{47.82}{19}{35} & \rank{69.24}{23}{35} & \rank{50.74}{20}{35} & \rank{73.96}{19}{35} & \rank{57.83}{20}{35} & \rank{65.45}{23}{35} & \rank{38.58}{22}{35} & \rank{64.97}{26}{35} & \rank{40.65}{25}{35} \\
UniDepth2-S &  & \rank{60.92}{29}{35} & \rank{39.78}{26}{35} & \rank{64.77}{26}{35} & \rank{47.02}{21}{35} & \rank{70.50}{22}{35} & \rank{45.88}{24}{35} & \rank{72.86}{26}{35} & \rank{53.65}{23}{35} & \rank{67.53}{9}{35} & \rank{44.70}{11}{35} & \rank{66.28}{15}{35} & \rank{45.11}{13}{35} \\
UniDepth2-S & \checkmark & \rank{60.87}{30}{35} & \rank{40.28}{25}{35} & \rank{64.61}{28}{35} & \rank{46.70}{22}{35} & \rank{70.72}{18}{35} & \rank{47.08}{23}{35} & \rank{72.60}{28}{35} & \rank{53.30}{24}{35} & \rank{65.60}{21}{35} & \rank{43.10}{13}{35} & \rank{66.19}{19}{35} & \rank{45.10}{14}{35} \\
DAv3-Mono-L & \checkmark & \rank{43.84}{31}{35} & \rank{23.94}{32}{35} & \rank{67.02}{11}{35} & \rank{53.09}{14}{35} & \rank{52.14}{31}{35} & \rank{27.94}{31}{35} & \rank{75.09}{12}{35} & \rank{63.39}{13}{35} & \rank{62.45}{27}{35} & \rank{34.43}{26}{35} & \rank{66.45}{13}{35} & \rank{41.81}{22}{35} \\
DAv3-Mono-L &  & \rank{43.84}{32}{35} & \rank{23.94}{31}{35} & \rank{66.96}{12}{35} & \rank{53.09}{15}{35} & \rank{52.14}{32}{35} & \rank{27.94}{31}{35} & \rank{75.10}{11}{35} & \rank{63.39}{12}{35} & \rank{62.45}{28}{35} & \rank{34.44}{25}{35} & \rank{66.26}{16}{35} & \rank{41.80}{23}{35} \\
DepthAnythingV2-S &  & \rank{40.35}{33}{35} & \rank{21.29}{35}{35} & \rank{62.15}{34}{35} & \rank{39.53}{35}{35} & \rank{49.99}{33}{35} & \rank{24.78}{34}{35} & \rank{71.49}{33}{35} & \rank{46.55}{33}{35} & \rank{56.71}{34}{35} & \rank{31.86}{35}{35} & \rank{67.09}{7}{35} & \rank{42.61}{20}{35} \\
DepthAnythingV2-B &  & \rank{39.80}{34}{35} & \rank{21.72}{34}{35} & \rank{62.57}{33}{35} & \rank{41.83}{32}{35} & \rank{48.36}{34}{35} & \rank{24.63}{35}{35} & \rank{71.54}{32}{35} & \rank{48.76}{32}{35} & \rank{57.13}{33}{35} & \rank{33.40}{28}{35} & \rank{66.97}{8}{35} & \rank{44.52}{16}{35} \\
DepthAnythingV2-L &  & \rank{17.83}{35}{35} & \rank{22.95}{33}{35} & \rank{56.69}{35}{35} & \rank{43.24}{29}{35} & \rank{14.18}{35}{35} & \rank{25.78}{33}{35} & \rank{65.00}{35}{35} & \rank{50.03}{29}{35} & \rank{36.77}{35}{35} & \rank{33.95}{27}{35} & \rank{64.53}{29}{35} & \rank{44.82}{15}{35} \\
\midrule GT &  & \rank{81.76}{0}{35} & \rank{89.18}{0}{35} & \rank{80.94}{0}{35} & \rank{88.59}{0}{35} & \rank{87.41}{0}{35} & \rank{93.00}{0}{35} & \rank{86.85}{0}{35} & \rank{92.55}{0}{35} & \rank{84.36}{0}{35} & \rank{92.02}{0}{35} & \rank{83.76}{0}{35} & \rank{91.38}{0}{35} \\
\midrule No Depth + \baselinecalib{} &  & \multicolumn{4}{c}{69.94} & \multicolumn{4}{c}{78.42} & \multicolumn{4}{c}{73.69} \\
\bottomrule
\end{tabular} 
        \\    
        ~
        \\
        \begin{tabular}{lcccccccccccc}
\toprule
  & \multicolumn{12}{c}{\mAA} \\ \cmidrule(lr){2-13}
  & \multicolumn{4}{c}{Mean} & \multicolumn{4}{c}{Statues} & \multicolumn{4}{c}{Vegetation} \\ \cmidrule(lr){2-5} \cmidrule(lr){6-9} \cmidrule(lr){10-13}
MDE-Backbone & \mysf{} & \sfro{} & \sfshift{} & \sfshiftro{} & \mysf{} & \sfro{} & \sfshift{} & \sfshiftro{} & \mysf{} & \sfro{} & \sfshift{} & \sfshiftro{}\\ \midrule
MoGeV1-L & \rank{62.48}{1}{19} & \rank{53.22}{1}{19} & \rank{44.17}{3}{19} & \rank{35.50}{3}{19} & \rank{70.82}{1}{19} & \rank{63.57}{1}{19} & \rank{52.29}{1}{19} & \rank{44.66}{1}{19} & \rank{60.14}{1}{19} & \rank{44.37}{2}{19} & \rank{42.75}{4}{19} & \rank{28.68}{3}{19} \\
Pi3-L & \rank{60.48}{2}{19} & \rank{49.78}{3}{19} & \rank{43.71}{4}{19} & \rank{35.00}{4}{19} & \rank{69.61}{2}{19} & \rank{61.34}{2}{19} & \rank{49.12}{6}{19} & \rank{40.34}{5}{19} & \rank{54.76}{9}{19} & \rank{35.64}{8}{19} & \rank{42.02}{6}{19} & \rank{24.60}{6}{19} \\
UniDepth2-L & \rank{60.47}{3}{19} & \rank{49.43}{4}{19} & \rank{44.29}{2}{19} & \rank{35.54}{2}{19} & \rank{67.68}{3}{19} & \rank{56.21}{5}{19} & \rank{50.82}{4}{19} & \rank{41.06}{4}{19} & \rank{58.77}{3}{19} & \rank{43.24}{3}{19} & \rank{44.01}{1}{19} & \rank{29.65}{2}{19} \\
UniK3D-L & \rank{59.89}{4}{19} & \rank{50.64}{2}{19} & \rank{44.92}{1}{19} & \rank{36.90}{1}{19} & \rank{66.69}{5}{19} & \rank{58.26}{4}{19} & \rank{51.69}{2}{19} & \rank{42.91}{3}{19} & \rank{60.08}{2}{19} & \rank{45.49}{1}{19} & \rank{43.83}{2}{19} & \rank{30.97}{1}{19} \\
MapAnything-G & \rank{58.79}{5}{19} & \rank{47.16}{6}{19} & \rank{39.51}{13}{19} & \rank{29.76}{8}{19} & \rank{66.03}{6}{19} & \rank{56.07}{6}{19} & \rank{48.39}{9}{19} & \rank{37.50}{7}{19} & \rank{57.11}{5}{19} & \rank{40.21}{5}{19} & \rank{36.98}{19}{19} & \rank{22.88}{10}{19} \\
MoGeV2-L & \rank{58.78}{6}{19} & \rank{48.05}{5}{19} & \rank{42.76}{5}{19} & \rank{33.37}{5}{19} & \rank{67.53}{4}{19} & \rank{60.82}{3}{19} & \rank{51.44}{3}{19} & \rank{43.03}{2}{19} & \rank{55.96}{8}{19} & \rank{36.26}{7}{19} & \rank{40.68}{12}{19} & \rank{25.80}{5}{19} \\
DepthPro-L & \rank{58.36}{7}{19} & \rank{44.03}{8}{19} & \rank{41.76}{8}{19} & \rank{29.58}{9}{19} & \rank{64.39}{9}{19} & \rank{52.11}{7}{19} & \rank{49.04}{7}{19} & \rank{35.57}{9}{19} & \rank{56.08}{7}{19} & \rank{33.88}{9}{19} & \rank{40.28}{13}{19} & \rank{21.51}{12}{19} \\
UniDepth2-B & \rank{58.23}{8}{19} & \rank{45.31}{7}{19} & \rank{42.74}{6}{19} & \rank{32.71}{6}{19} & \rank{64.99}{7}{19} & \rank{50.97}{8}{19} & \rank{48.82}{8}{19} & \rank{35.83}{8}{19} & \rank{57.85}{4}{19} & \rank{40.41}{4}{19} & \rank{42.39}{5}{19} & \rank{27.75}{4}{19} \\
UniDepth1-L & \rank{55.75}{9}{19} & \rank{36.80}{13}{19} & \rank{39.78}{11}{19} & \rank{22.92}{14}{19} & \rank{62.37}{11}{19} & \rank{44.87}{11}{19} & \rank{46.15}{12}{19} & \rank{27.22}{14}{19} & \rank{54.68}{10}{19} & \rank{31.91}{11}{19} & \rank{40.83}{10}{19} & \rank{20.87}{14}{19} \\
Metric3DV2-G & \rank{55.64}{10}{19} & \rank{38.70}{10}{19} & \rank{40.11}{10}{19} & \rank{24.69}{12}{19} & \rank{64.68}{8}{19} & \rank{47.58}{10}{19} & \rank{46.58}{11}{19} & \rank{29.61}{12}{19} & \rank{49.81}{15}{19} & \rank{27.02}{17}{19} & \rank{39.68}{14}{19} & \rank{18.20}{18}{19} \\
DAv3-Metric-L & \rank{54.85}{11}{19} & \rank{39.41}{9}{19} & \rank{39.70}{12}{19} & \rank{26.08}{11}{19} & \rank{62.71}{10}{19} & \rank{48.79}{9}{19} & \rank{47.62}{10}{19} & \rank{34.01}{10}{19} & \rank{53.78}{12}{19} & \rank{32.90}{10}{19} & \rank{37.65}{18}{19} & \rank{19.98}{15}{19} \\
Metric3DV2-L & \rank{54.54}{12}{19} & \rank{37.13}{12}{19} & \rank{39.42}{14}{19} & \rank{24.62}{13}{19} & \rank{60.70}{13}{19} & \rank{42.63}{13}{19} & \rank{45.45}{14}{19} & \rank{28.26}{13}{19} & \rank{49.39}{16}{19} & \rank{26.43}{18}{19} & \rank{38.74}{16}{19} & \rank{19.35}{16}{19} \\
Metric3DV2-S & \rank{53.37}{13}{19} & \rank{33.53}{14}{19} & \rank{38.63}{16}{19} & \rank{21.15}{17}{19} & \rank{58.82}{14}{19} & \rank{39.68}{14}{19} & \rank{44.15}{16}{19} & \rank{24.77}{16}{19} & \rank{50.98}{14}{19} & \rank{24.58}{19}{19} & \rank{37.95}{17}{19} & \rank{15.56}{19}{19} \\
UniDepth2-S & \rank{53.29}{14}{19} & \rank{37.41}{11}{19} & \rank{40.21}{9}{19} & \rank{27.31}{10}{19} & \rank{62.15}{12}{19} & \rank{44.23}{12}{19} & \rank{45.69}{13}{19} & \rank{31.03}{11}{19} & \rank{56.78}{6}{19} & \rank{36.62}{6}{19} & \rank{42.01}{7}{19} & \rank{24.30}{7}{19} \\
UniDepth1-C & \rank{53.19}{15}{19} & \rank{31.34}{15}{19} & \rank{38.94}{15}{19} & \rank{19.81}{18}{19} & \rank{57.44}{15}{19} & \rank{35.83}{15}{19} & \rank{44.52}{15}{19} & \rank{22.17}{19}{19} & \rank{54.34}{11}{19} & \rank{29.34}{15}{19} & \rank{40.76}{11}{19} & \rank{18.74}{17}{19} \\
DAv3-Mono-L & \rank{39.64}{16}{19} & \rank{26.98}{16}{19} & \rank{41.96}{7}{19} & \rank{30.93}{7}{19} & \rank{46.57}{16}{19} & \rank{32.94}{16}{19} & \rank{50.04}{5}{19} & \rank{39.45}{6}{19} & \rank{52.67}{13}{19} & \rank{30.30}{12}{19} & \rank{39.49}{15}{19} & \rank{22.06}{11}{19} \\
DepthAnythingV2-L & \rank{37.58}{17}{19} & \rank{26.66}{17}{19} & \rank{37.28}{18}{19} & \rank{22.82}{15}{19} & \rank{43.72}{17}{19} & \rank{31.37}{17}{19} & \rank{43.21}{18}{19} & \rank{25.66}{15}{19} & \rank{47.08}{17}{19} & \rank{29.65}{13}{19} & \rank{41.74}{8}{19} & \rank{23.70}{8}{19} \\
DepthAnythingV2-B & \rank{36.36}{18}{19} & \rank{25.21}{18}{19} & \rank{37.72}{17}{19} & \rank{22.11}{16}{19} & \rank{42.58}{19}{19} & \rank{30.33}{18}{19} & \rank{43.87}{17}{19} & \rank{24.72}{17}{19} & \rank{47.01}{18}{19} & \rank{29.38}{14}{19} & \rank{42.96}{3}{19} & \rank{23.47}{9}{19} \\
DepthAnythingV2-S & \rank{35.79}{19}{19} & \rank{23.97}{19}{19} & \rank{36.48}{19}{19} & \rank{19.76}{19}{19} & \rank{43.10}{18}{19} & \rank{29.30}{19}{19} & \rank{43.20}{19}{19} & \rank{22.23}{18}{19} & \rank{45.36}{19}{19} & \rank{27.35}{16}{19} & \rank{41.33}{9}{19} & \rank{21.38}{13}{19} \\
\midrule GT & \rank{80.17}{0}{19} & \rank{88.52}{0}{19} & \rank{62.12}{0}{19} & \rank{79.66}{0}{19} & \rank{86.46}{0}{19} & \rank{92.47}{0}{19} & \rank{67.91}{0}{19} & \rank{84.01}{0}{19} & \rank{82.48}{0}{19} & \rank{91.08}{0}{19} & \rank{67.64}{0}{19} & \rank{85.21}{0}{19} \\
\midrule No Depth + \baselinesf{} & \multicolumn{4}{c}{58.59} & \multicolumn{4}{c}{66.45} & \multicolumn{4}{c}{63.59} \\
\bottomrule
\end{tabular}
        \end{tabular}
    }
    
    \label{tab:sm_d2p_splg}
\end{table}

\section{Computational Requirements}
\label{sec:appendix_computational_requirements}
To run all experiments on the D2P benchmark we utilized 13 GPU hours on a cluster with A100 40 GB GPUs mostly for MDE inference and matching and 200 CPU hours of Intel Xeon Gold 6338 2.00 GHz CPUs. For the experiments on the standard benchamark we have used 30 GPU hours and 600 CPU hours.

\section{Broader Impact}
\label{sec:appendix_broader_impact}
This work proposes a new evaluation framework and dataset for monocular depth estimation, which can improve the assessment of depth models in real-world conditions. A potential positive impact is enabling more reliable deployment of systems that rely on depth estimation, such as robotics, autonomous navigation, and augmented reality, by promoting evaluation protocols that better reflect downstream performance.
Potential negative impacts arise from improved evaluation and development of depth estimation models, which may indirectly benefit applications such as surveillance, tracking, or military systems. In addition, models evaluated using our framework may still fail in challenging conditions, and overreliance on such systems without proper validation could lead to safety risks in real-world deployments.
Our work does not introduce new modeling capabilities but rather focuses on evaluation. Nevertheless, it may contribute to safer deployment by encouraging more realistic, task-driven evaluation protocols that help identify failure modes and improve robustness.

%%%%%%%%%%%%%%%%%%%%%%%%%%%%%%%%%%%%%%%%%%%%%%%%%%%%%%%%%%%%

\end{document}